\documentclass[preprint]{elsarticle}

\usepackage{amssymb}
\usepackage{lipsum}
\usepackage[dvipsnames]{xcolor}
\usepackage{svg}
\usepackage{amsmath} 
\usepackage{bm}
\usepackage{marginnote}
\usepackage{float}
\usepackage{ulem}
\usepackage{graphicx}
\usepackage{subcaption}
\usepackage[export]{adjustbox}% http://ctan.org/pkg/adjustbox
\usepackage{acronym}
\usepackage{lineno}
\journal{ }

\begin{document}

\begin{frontmatter}

\title{Can physical information aid the generalization ability of Neural Networks for hydraulic modeling?}


\author[first]{Gianmarco Guglielmo}
\author[first]{Andrea Montessori}
\author[first,second]{Jean-Michel Tucny}
\affiliation[first]{organization={Roma Tre University},%Department and Organization
            addressline={Department of Civil, Computer Science, and Aeronautical Technologies Engineering, Via Vito Volterra 62, 00146}, 
            city={Rome},
            country={Italy}}

\author[first]{Michele La Rocca}
\author[first]{Pietro Prestininzi}
\affiliation[second]{organization={Center for Life Nano- \& Neuro-Science, Fondazione Istituto Italiano di Tecnologia (IIT)},%Department and Organization
            addressline={Viale Regina Elena 295, 00161}, 
            city={Rome},
            country={Italy}} 

\begin{abstract}
Application of Neural Networks to river hydraulics is fledgling, despite the field suffering from data scarcity, a challenge for machine learning techniques. Consequently, many purely data-driven Neural Networks proved to lack predictive capabilities. In this work, we propose to mitigate such problem by introducing physical information into the training phase. The idea is borrowed from Physics-Informed Neural Networks which have been recently proposed in other contexts. Physics-Informed Neural Networks embed physical information in the form of the residual of the Partial Differential Equations (PDEs) governing the phenomenon and, as such, are conceived as neural solvers, i.e. an alternative to traditional numerical solvers.
Such approach is seldom suitable for environmental hydraulics, where epistemic uncertainties are large, and computing residuals of PDEs exhibits difficulties similar to those faced by classical numerical methods. 
Instead, we envisaged the employment of Neural Networks as neural operators, featuring physical constraints formulated without resorting to PDEs. The proposed novel methodology shares similarities with data augmentation and regularization.
We show that incorporating such \textit{soft} physical information can improve predictive capabilities.
\end{abstract}



\end{frontmatter}


\section{Introduction}
\label{introduction}

As common to environmental hydraulics, river flood mapping requires models able to explore scenarios not entirely encompassed by the set of available observations, i.e. featuring predictive/generalization capabilities. An illustrative example is the simulation of high hydraulic hazard scenarios by means of models calibrated only on observations of medium to low flow regimes \cite{di2009uncertainty}. 

Recently, the need for reliable flood maps of ungauged basins has become urgent \cite{bloschl2013runoff, cole2006issues, hrachowitz2013decade}, the latter ranging from small catchments to scarcely populated large regions in developing countries. In addition to the difficulties related to the lack of high quality and/or scarce measurements, mapping vast flood-prone areas by means of physically based models requires a considerable computational burden, even assuming drastic conceptual simplifications \cite{prestininzi2011selecting}. Classical modeling approaches \cite{kumar2023comprehensive} may thus become unfeasible in such cases, and resorting to the exploitation of similarities with other basins has been envisaged. Indeed, Machine Learning (ML) has been proposed \cite{dasgupta2023revisit} as a fruitful way to overcome the above issues, with Neural Networks (NNs) specifically employed in the context of environmental hydraulics \cite{kratzert2019toward}. A comprehensive review has been developed by \cite{bentivoglio2022deep}. An example of prediction of urban pluvial flood by means of Convolutional Neural Networks (CNNs) is given by \cite{lowe2021u}, while a CNN for fluvial flood inundation is developed by \cite{kabir2020deep}. 

However, the above referenced first attempts seem to point out a substantial inability of the trained NNs to generalize, that is to provide a reasonable prediction for scenarios even slightly different from the ones over which they were trained on \cite{nguyen2023integration}. The main reason of such flaw is the lack of sufficiently large training datasets \cite{guo2021data}, resulting in overfitted models. An effect closely related to overfitting and, as such, ascribable to the scarcity of calibration data, is the frequent violation of conservation laws \cite{berkhahn2019ensemble}, resulting in marked non-physical results. 

A recent new paradigm is represented by Physics‐Informed Machine Learning (PIML) \cite{karniadakis2021physics}. PIML consists of augmenting the informational content of the training datasets by introducing physically‐based constraints into existing ML models. By doing so, it is theoretically possible
to reduce the machine's learning time, increase its generalization capabilities \cite{jamali2021machine}, and obtain physically‐based results. The physical constraints typically require the machine output to comply with conservation equations \cite{jagtap2020conservative}.

The widespread approach of PIML generated a new class of NNs, namely the Physics-Informed Neural Networks (PINNs), introduced by \cite{raissi2019physics} and intended as \textit{neural solvers} \cite{hao2022physics}, i.e. a NN specifically designed to find the behavior of a unique physical phenomenon described by a known differential problem. PINNs allow for solving both forward and inverse problems governed by a known partial differential equation (PDE), initial and boundary conditions \cite{lu2021deepxde}. They typically modify the loss function by adding a term containing the residual of the PDE.  
Most of the highly specialized algorithms proposed for computing the derivatives of PDEs rely on Automatic Differentiation \cite{baydin2018automatic}, although Numerical Differentiation or a combination of both approaches \cite{chiumarrying} have also been considered. 

Recently, PINNs have been applied to a wide range of benchmark problems of fluid mechanics \cite{cai2021physics}, in particular as neural solvers for Navier-Stokes equations \cite{jin2021nsfnets}, and have shown the potential to become more practical than classical techniques in many circumstances \cite{cuomo2022scientific}. The advantages of such approach, compared to numerical models, are certainly the possibility of seamlessly integrating empirical data into the model and the ability to address problems that are mathematically ill-posed.
Additionally, it enables tackling issues whose dimensionality is so high that the use of traditional solvers would result in dramatically elevated computational burden.

However, we believe employing PINNs as neural solvers is hardly applicable to river hydraulics. Indeed, PINNs are designed to approximate the solution to a PDE within a specific domain, and every change requires training a new PINN. Furthermore, what is being sought is not a surrogate for the numerical model solving for a single instance of a physical system, which would be subject to the same, if not additional, limitations and difficulties previously described. Instead, the aim is to design a \textit{neural operator} \cite{hao2022physics} that can effectively approximate an entire class of physical problems. 

In the field of flood simulations, the challenge for ML applications is to employ scarce observations for the training phase while being able to generalize reliably over new scenarios. Therefore, in this work we assess whether the performance of a deep learning model can be enhanced by incorporating \textit{soft} physical information, i.e. without resorting to the calculation of the residual of a differential equation (\textit{hard} physical information). 
This feature can be a significant advantage in applications characterized by a large epistemic uncertainty, i.e. the conceptual model of the phenomenon is partially or entirely unknown. In the context of flood simulations, where hydraulically relevant distributed parameters such as roughness, lithology, topography etc. pose significant uncertainties, introducing \textit{hard} physical information like either the residual of the 2D Shallow-Water equations \cite{qian2019physics} or the 1D Saint-Venant equations \cite{mahesh2022physics}, may not be advisable. 

In this work, we extend the concept of PINNs with respect to two fundamental points: 

- the NN is not used to obtain a tailored solution of a single problem (neural solver), instead the aim is to construct a neural operator capable of capturing the relationships between input and output functions across an entire class of physical phenomena (neural operator);

- the insertion of physical information does not necessarily have to be linked to the residual of the differential equation (hard physical information), but can involve other physical properties of the system and the data (soft physical information). This method can be interpreted as a special form of physical data augmentation, involving both inputs and outputs. 

We then assessed the gain in generalization capability of such physical aided neural operators compared to the purely Data-Driven (DD) ones in the context of environmental hydraulics.

The problem solved by our NNs consists of a simple yet non-trivial physical problem, namely the reconstruction of a steady state, one-dimensional, water surface profile in a rectangular channel. This problem was investigated by \cite{cedillo2022physics} for steady flow using PINNs as neural solvers.
Its simplicity stems from both the low dimensionality of the feature space, which allows for the adoption of small NNs, and the possibility to easily generate dataset of exact solutions at will. Moreover, the choice of an idealized case over a more realistic one mitigates the risk of producing biased results as a consequence of possible prevailing aspects of the specific case, thus preventing the effects of the proposed modifications to be properly assessed.
The highly informative content of the analyzed problem comes mainly from the hidden complexity of the underlying physics, associated to the possible occurrence of a hydraulic jump, a phenomenon which does not belong to the set of solutions of the underlying differential equations. 

The paper is structured as follows: section Methodology contains the overall plan followed in the work; the Numerical experiments section details the structures of the NNs and describes how the soft physical information is employed; then the Discussion of results follows, defining the assessment procedures. Conclusions are then drawn and perspectives are advanced. 

\section{Methodology}
\label{Methodology}

\subsection{Data-Driven and Physics-Informed Neural Networks}
Neural Networks \cite{lecun2015deep} aim to capture (i.e. learn) complex mappings between inputs $\mathbf{x}$ and outputs $\mathbf{\hat y} $, described by an unknown non-linear function $G^*$:
\begin{equation}
G^*(\mathbf{ x}) = \mathbf{\hat y} 
\label{shallowNN - true}
\end{equation}
By tuning the values of the parameter vector $\boldsymbol{\theta}$, the NN learns to mimic $G^*$ through the approximate function $G$:
\begin{equation}
 G(\mathbf{x}; \boldsymbol{\theta}) = \mathbf{y} 
\label{shallowNN2}
\end{equation}
where $\mathbf{y}$ are the predicted values.

The NN parameters are learned from data by leveraging information derived from the training set, namely the couples $(\mathbf{x}, \mathbf{\hat y})$. In a purely DD training process, the network calibrates its parameters by minimizing a loss function  $\mathcal{L}$ between $\mathbf{y}$ and $\mathbf{\hat y}$: 
\begin{equation}
arg \ min_{\boldsymbol{\theta}}\mathcal{L} (\mathbf{y},\mathbf{\hat y})
\label{minimization problem}
\end{equation}
A common choice for the loss function in a DD model for a regression problem (the output variables are real unbounded numbers) is the Mean Squared Error (MSE):
\begin{equation}
MSE = \frac{\sum_{i=1}^{N} (y_i -\hat{y_i})^2}{N}
\label{MSE}
\end{equation}
but other metrics can be used depending on the nature of the data and the goals of the model. For further insights into Feed-Forward Neural Networks (FFNNs) and their training process, additional information can be found in \ref{FFNN RNN}.

PINNs are a new paradigm of ML models that combine principles from physics-based modeling with the flexibility of NNs. The key concept is to embed well-known physical laws into the training phase of a NN, integrating data with mathematical models.

PINNs have been used as neural solvers for physical systems governed by a known set of Partial Differential Equations (PDEs), addressing both forward problems (solving the equation) and inverse problems (determining unknown parameters). PINNs can even handle situations where the problem is mathematically ill-posed and therefore classical numerical methods are unfeasible. 

The strategy pursued by PINNs lies in building the loss function as follows:
\begin{equation}
\mathcal{L} = \lambda \cdot \mathcal{L}_{DD}(\mathbf{y}, \mathbf{\hat y}) + (1- \lambda) \cdot \mathcal{L}_{P}(\mathbf{x}, \mathbf{y})
\label{PINN loss}
\end{equation}
where $\mathcal{L}_{DD}$ is a metric which measures the distance between predicted $\mathbf{y}$ and observed data $\mathbf{\hat y}$ (e.g. MSE, equation \eqref{MSE}), $\mathcal{L}_{P}$ measures the residual of the PDE sampled at a set of points within the PDE domain, $\lambda$ balances the data and physical contributions, ranging between 0 and 1.

\subsection{Soft physical information}
\label{extension}
As for the PINNs, the proposed inclusion of soft physical information in the training phase of a neural operator is achieved through an additional physical term $\mathcal{L}_{P}$ in the loss function, as shown in equation \eqref{PINN loss}. Various physical principles can be encoded into this term, enabling the model to capture a broader spectrum of physical behaviors and relationships hidden in the data. 

While the term $\mathcal{L}_{DD}$ still depends solely on the predicted outputs $\mathbf{y}$ and the true outputs $\mathbf{\hat y}$, the term $\mathcal{L}_{P}$ is a metric that can also involve the inputs $\mathbf{x}$. Therefore, the loss function $\mathcal{L}$ incorporating soft physical information reads:
\begin{equation}
\mathcal{L} = \lambda \cdot \mathcal{L}_{DD}(\mathbf{y}, \mathbf{\hat y}) + (1- \lambda) \cdot \mathcal{L}_{P}(\mathbf{x}, \mathbf{y}, \mathbf{\hat y} )
\label{PINN loss soft}
\end{equation}

This approach shares similarities with two techniques, namely the introduction of a regularization term and data augmentation. 

The former limits the growth of weights during the training phase, thus mitigating possible overfitting effects \cite{ng2004feature}. In this sense, the physical term $\mathcal{L}_{P}$ acts as a regularization term.

Data augmentation \cite{shorten2019survey} is a technique used to artificially increase the size of the training dataset by applying various transformations to the input data. The goal is to enhance the model's performance by exposing it to a more diverse set of examples, which helps improving generalization and robustness.
Our approach allows for an enrichment of the informational content of the dataset by computing new physically-based quantities from input data $\mathbf{x}$ and the predicted outputs $\mathbf{y}$. The $\mathcal{L}_{P}$ term compares such quantities with their true values, which depend on input data $\mathbf{x}$ and the true outputs $\mathbf{\hat y}$. From this perspective, this approach can be considered as a kind of physically-based data augmentation and it allows for enriching the informational capacity of the data. 

As this approach involves only the loss function, it is a priori applicable to a wide range of physical systems and compatible with all NN architectures.

\subsection{Synthetic case definition}
\begin{figure}[H]
	\centering 
		\includegraphics[width=1.\textwidth, angle=0]{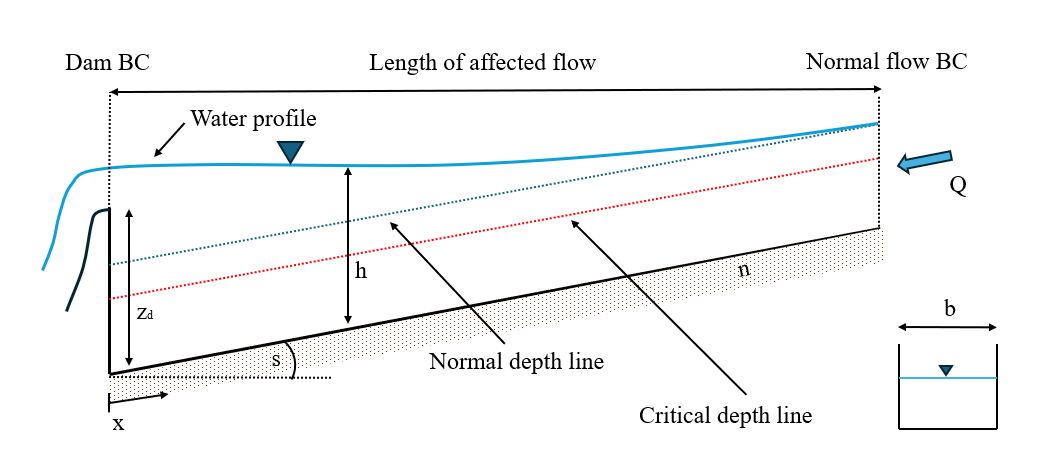}
		\includegraphics[width=1.\textwidth, angle=0]{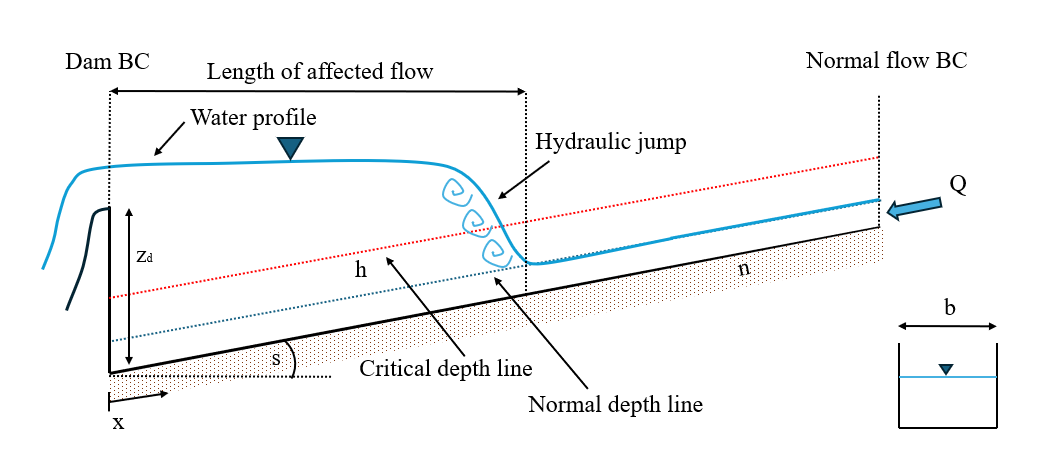}	
	\caption{Sketch of the analyzed physical problem, i.e. the steady state water profile in a rectangular prismatic channel. Upper panel describes the subcritical case; lower panel depicts the mixed regime case. For the meaning of the symbols, the reader is referred to the text.}
	\label{fig:sketch}
\end{figure}

We construct a specific synthetic case study to test and show the proposed methodology. 
We assess the performance of several approaches in the reconstruction of the water surface profile along a 1D channel induced by the presence of a weir placed at the outlet cross section. Such problem mimics a commonly occurring scenario of determining the area affected by the presence of an inline structure in a river. Additionally, if a supercritical flow regimes develops in the upstream part of the channel, a hydraulic jump occurs (Figure \ref{fig:sketch}).
The water profile encompassing a hydraulic jump does not belong to the set of solutions of the differential equation generating the gradually-varied water profile. Indeed, it is an internal boundary whose location (and strength of the local discontinuity) needs to be solved for through additional information (see \ref{dataset generation}). Therefore, approaching the problem within a classical PINN framework (namely the neural solver) would be non-trivial, since the jump condition would require a tailored mathematical description. 

The solution of the physical problem, assuming a cylindrical rectangular channel and steady flow conditions, is a function $F^*$ which maps from \(\mathbb{R}^6\) to \(\mathbb{R}\):
\begin{equation}
\hat{h} = F^*(x, s, b, n, z_d, Q)
\label{problem}
\end{equation}
with $\hat{h}$ representing the true flow depth;
$x$, the distance from the dam; 
$s$, the channel slope;
$b$, the channel width;
$n$, the Manning coefficient (related to the channel's roughness);
$z_d$, the height of the dam; 
and $Q$, the water discharge.

The gradual varied flow assumption rules out the presence of discontinuities in the depth profile. However, the transition from supercritical to subcritical flow occurs through the appearance of a hydraulic jump which needs to be accounted for as an internal boundary whose location and strength need to be solved for through additional information.

\section{Numerical experiments}
\label{Numerical experiments}
\subsection{Generation of the training dataset}
A set of 10390 profiles were generated by uniformly varying $s$, $b$, $n$, $z_d$, and $Q$, solving equation \ref{energy equation} using a Finite-Difference scheme, with a fixed spatial discretization of $\Delta x$ set to 10 meters, covering a total length of 5000 meters. This results in profiles composed of 501 data points.

Therefore, we can interpret the problem as approximating $F^*$ in equation \eqref{problem} through a NN, having at our disposal a uniform sampling of the six-dimensional function's domain.

The obtained profiles were then randomly divided into training profiles (70 \%, totaling 7274 profiles), while the remaining ones constitute the validation set and the test set (both consisting of 1558 profiles).

The input data to the various models are always normalized using the Standard Scaler, which is a commonly used data pre-processing technique to scale the features to have a mean of zero and a standard deviation of one. This is necessary to prevent troublesome issues during training due to differences in the numerical values of the features \cite{ali2014data}. 

\subsection{The three architectures}
Three FFNN architectures were examined, namely the Single-Point (SP), the INTegrator (INT), and the Vector-To-Sequence (VTS). We would like to point out that the aim of this work is not to select the best architecture for solving the problem at hand, but to evaluate the effects of a physically informed training.
Moreover, while it was also possible to employ Recurrent Neural Networks, interpreting the temporal series output as the spatial series of the water depth (namely the water profiles), we chose to limit the analysis to FFNNs, aligning with the FFNN architecture of classical PINNs. 

\begin{figure}[H]
	\centering 
		\includegraphics[width=1.\textwidth, angle=0]{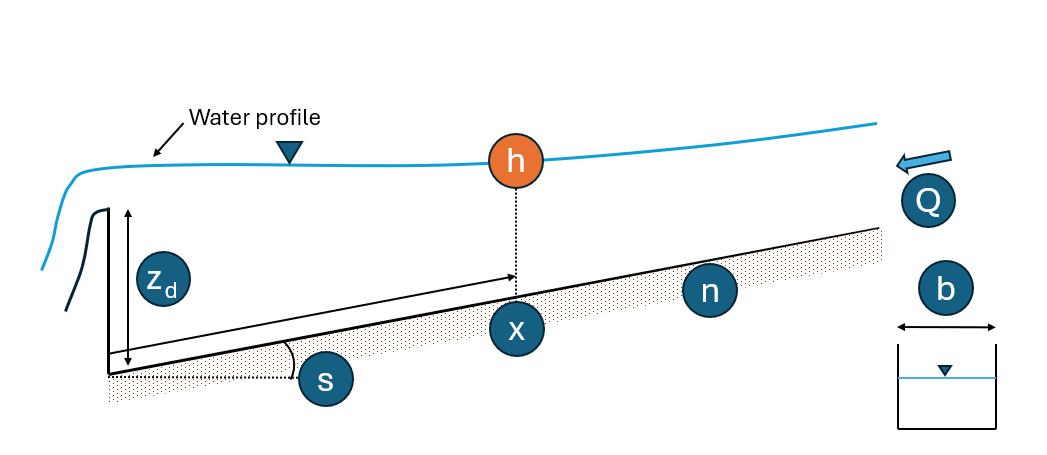}
		\includegraphics[width=1.\textwidth, angle=0]{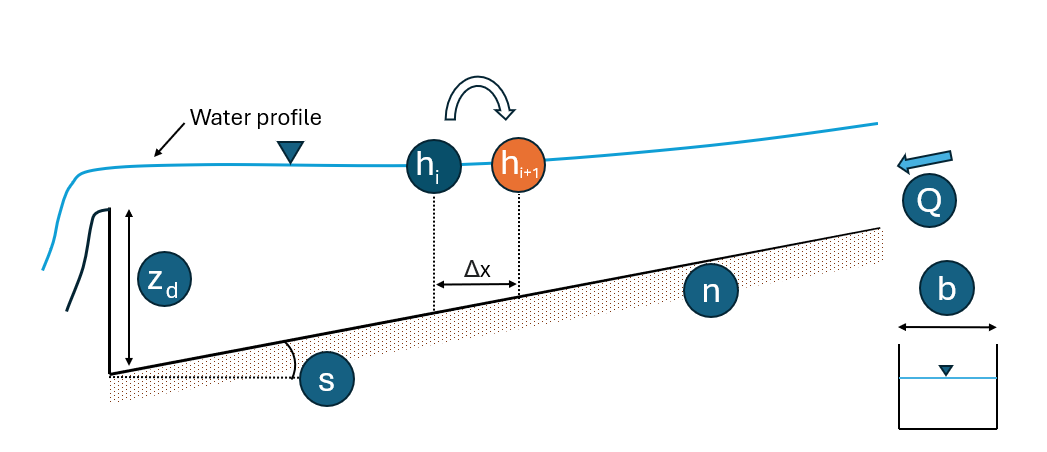}
		\includegraphics[width=1.\textwidth, angle=0]{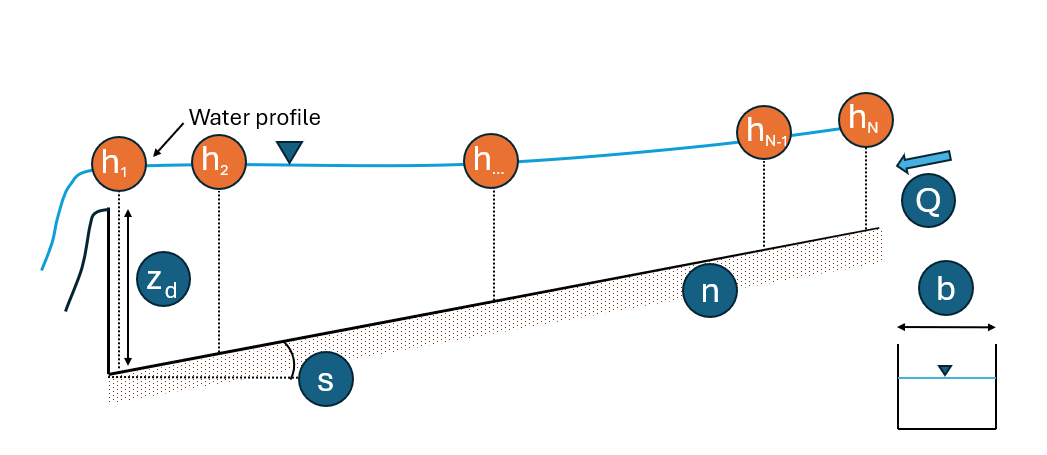}
	\caption{Profile reconstruction using the three different approaches; input and output are depicted as blue and orange circles, respectively; Single-Point (upper panel) outputs the flow depth at a specific stationing; Integrator (middle panel) requires the neighboring downstream value, regardless of the stationing; Vector-To-Sequence outputs the entire vector of flow depth (i.e. the whole water profile) at once. }
	\label{fig:reconstruction}
\end{figure}

\begin{figure}
		\centering
		\includegraphics[width=0.4\textwidth,angle=-90,valign=c]{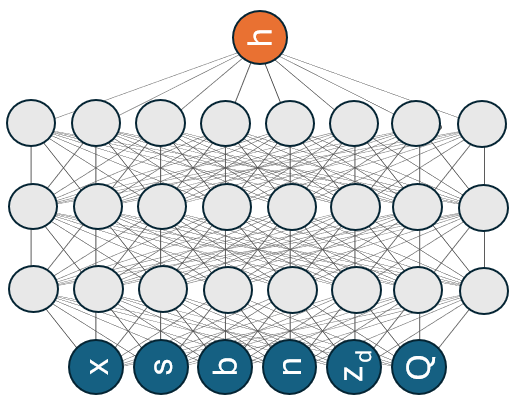}
		\includegraphics[width=0.4\textwidth,angle=-90,valign=c]{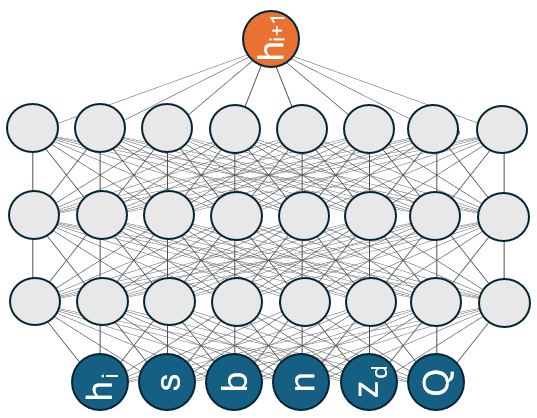}
		\includegraphics[width=0.6\textwidth,angle=-90,valign=c]{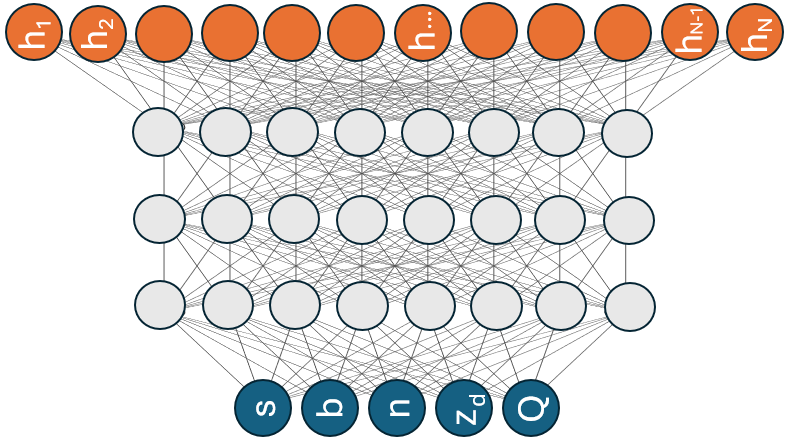}
	\caption{The employed FFNN architectures. Left to right: Single-Point, Integrator, Vector-To-Sequence.}
	\label{fig:architectures}
\end{figure}

A sketch of the operating principles of SP, INT, and VTS architectures is illustrated in Figure \ref{fig:reconstruction} and the corresponding network topologies are depicted in Figure \ref{fig:architectures}. Further details are in order. 

\subsubsection{Single-Point}
A verbatim translation of the problem formulated in Eq. \eqref{problem} is to employ the NN to approximate the function $F^*$ with $F$:
\begin{equation}
h = F(x, s, b, n, z_d, Q; \boldsymbol{\theta})
\label{SP1}
\end{equation}

In this approach, the FFNN takes the six quantities governing the
phenomenon as inputs and predicts a single value for the water depth $h$ at the stationing $x$ as an output.

To reconstruct the whole profile  $\mathbf{h}$, a \textit{SP} model needs to be run with all the desired stationing values.

\subsubsection{Integrator}
In this method, a NN is utilized in the guise of a numerical integrator, that is aimed at determining the local water depth based on its value at the adjacent stationing. An eigenanalysis of the differential problem underpinning a steady-state 1D free surface flow like the one chosen in this study, would require to adopt either a downstream or upstream oriented solution direction based on the local flow regime, namely super or subcritical, respectively. However, in the context of surrogate models like the one based on ML, due to the lack of any physical support, such requirement can be overlooked. In the following, an upstream moving algorithm has been chosen. 
This methodology requires a rearrangement of the dataset into the pairs
$\left([\hat{h}_i, s, b, n, z_d, Q], \hat{h}_{i+1}\right)$,
where $\hat{h}_i$ and $\hat{h}_{i+1}$ must be uniformly spaced by $\Delta x$, where $i$ spans all stationing points of all profiles. 

Due to the above structure, the INT approach misses any information regarding the distance from the dam.  As a result, unlike the SP, it cannot predict the height at any distance from the dam but only at multiples of the fixed $\Delta x$ set by the dataset.

\begin{equation}
h_{i+1} = INT(h_i, s, b, n, z_d, Q; \boldsymbol{\theta})
\label{int}
\end{equation}

Starting from a known downstream boundary condition ($h_1$), the integrator $INT$ can be applied recursively. At each stationing, the output from the downstream one serves as input, eventually leading to the reconstruction of the whole profile.
\begin{equation}
h_i = INT(h_{i-1},...) \circ INT(h_{i-2}, ...)\circ ... \circ INT(h_1, s, b, n, z_d, Q; \boldsymbol{\theta})
\label{int method}
\end{equation}

It is essential to note that any classical numerical integrator would require a repeated check for the occurrence of a hydraulic jump, as well as an \textit{ad hoc} procedure for its solution. Instead, this model can be applied flawlessly across such discontinuity.

Two further specific features of the Integrator approach are in order. Firstly, at each location, due to its recursive application, \textit{INT} outputs a depth value whose error depends on the accuracy of the previous applications. As a consequence, the model's accuracy generally decreases in the marching direction of the algorithm. A side advantage of such upstream marching algorithm lies in the possibility to incorporate physical conservation balances between the current and previous location. Secondly, the recursive application of \textit{INT} introduces a subtle advantage over the other purely DD approaches: indeed its initialization (occurring at the most downstream stationing in our case) represents an implicit imposition of a physical constraint.

\subsubsection{Vector-To-Sequence}
The Vector-To-Sequence employs a FFNN (namely $V$) that receives as input the five parameters determining the profile solution and predicts as output the entire vector $\mathbf{h}$.

\begin{equation}
\mathbf{h} = V(s, b, n, z_d, Q; \boldsymbol{\theta})
\label{VtS}
\end{equation}

The dataset must be rearranged in pairs $\left([ s, b, n, z_d, Q], \hat{\mathbf{h}}\right)$. Just like for the INT approach, the input does not include stationing data, thus implying that the spacing of the output matches the one of the training dataset.

An advantage of this architecture is that it allows for the implementation of physical loss terms whose formulation requires the knowledge of the whole profile, e.g. the volume of water.

\subsection{Physical training strategies}

Each of the abovementioned three NN architectures has undergone both a purely DD and a physically informed training, the latter consisting in exploiting the local values of the: 

\begin{itemize}
    \item Specific energy, as in equation \eqref{specific energy}, EN strategy in the following; 
    \item Froude number, as in equation \eqref{Froude}, FR strategy in the following. 
\end{itemize}
Since the VTS approach outputs the entire profile at once, it was also possible to test three additional training strategies using: 
\begin{itemize}
\item The volume of water flowing in the river (namely, the area under the water profile), VOL strategy in the following;

\item  The downstream boundary condition, BC strategy in the following;

\item  The energy differential equation, PDE strategy in the following.
\end{itemize}

The loss function for the purely DD training strategy is formulated in terms of MSE between real and predicted water depth. For the physics-informed training strategies, this function is augmented with an additional loss term, as shown in Equation \eqref{PINN loss soft}. 

The physical loss terms adopted for each of the above list of training strategies, read:

\begin{equation}
\mathcal{L}_P^{EN} =  \frac{\sum_{i=1}^{N} (E(h_i) - E(\hat{h}_i))^2}{N}
\label{Loss Energy}
\end{equation}
\begin{equation}
\mathcal{L}_P^{FR}  =  \frac{\sum_{i=1}^{N} (Fr(h_i) -Fr(\hat{h}_i))^2}{N}
\label{Loss Froude}
\end{equation}
\begin{equation}
\mathcal{L}_P^{VOL}  =  \sum_{i=1}^{N}h_i-\sum_{i=1}^{N}\hat{h}_i
\label{Loss Volume}
\end{equation}
\begin{equation}
\mathcal{L}_P^{BC}  =  |h_1 - \hat{h}_1|
\label{Loss BC}
\end{equation}
\begin{equation}
\mathcal{L}_P^{PDE}  = \sum_{i=2}^{N-1} \left( \frac{E_{i+1}(h_i)-E_{i-1}(h_i)}{2 \Delta x} + i - J_i(h_i) \right)
\label{Loss PDE FD}
\end{equation}

The first four constraints above are what we refer to as \textit{soft} physical information. The last constraint, which is a \textit{hard} physical information, is only formally borrowed from the classical PINN paradigm: indeed it is here employed in the context of a neural operator, i.e. aimed at improving its training phase. 
In equations \eqref{Loss Energy} - \eqref{Loss Froude},
$N$ is to be considered as the batch size, that is the number of training examples utilised at each training iteration, for the SP and INT architectures (which have point data), while for VTS, $N$ in equations \eqref{Loss Energy} - \eqref{Loss PDE FD} represents the length of the data series. Formally, to obtain the loss function at each iteration, for the VTS, the average of the values obtained over the batch size must be considered (though this is not shown here to avoid burdening the notation). 

As introduced in Section \ref{extension}, it is worth recalling that the $\mathcal{L}_P$ term acts as a regularization term, and the physical training strategies can be interpreted as a form of physical data augmentation. Hereinafter, the term "model" refers to the combination of an architecture and a training strategy.

\subsection{Hyperparameters}
Hyperparameters are values to be set before the training process and that are not updated during the training phase. They encompass crucial features such as the number of hidden layers and neurons, the optimization algorithm along with its learning rate, and parameters related to early stopping. For the SP, we initially assume 3 layers with 30 neurons each. Similarly, for the INT, we also use 3 layers with 30 neurons. As for the VTS, we opt for 3 layers with 40 neurons each, considering the higher number of outputs the network needs to produce compared to the other architectures. The ReLU activation function \cite{nair2010rectified} is consistently employed for each neuron. The number of layers is fixed for all NNs as the analysis of its influence is beyond the scope of this work.

We implemented each neural network using TensorFlow \cite{tensorflow2015-whitepaper} and Keras \cite{chollet2015keras}. We employ a learning rate reduction technique, specifically ReduceLROnPlateau, within the Adam \cite{kingma2014adam} optimization algorithm. The initial learning rate is set to $0.001$ and is progressively reduced when approaching a minimum of the loss function. We also employ an early stopping criterion during the training phase, based on the MSE calculated on the validation set. All these hyperparameters are fixed within each model to achieve consistent results. 

However, it is important to recall here that the focus of this work is not to evaluate the best model to solve the issue at hand, but rather the assessment of the effects of the proposed methodology within each architecture. Thus any influence of the topology that might favor one architecture over the other can be overlooked. When adopting the physical training strategy, the value for the hyperparameter $\lambda$ in equation \eqref{PINN loss} has been chosen as the one yielding the best performance in the interval $[0,1]$.

\section{Discussion of results}
\label{Results}
\begin{figure}
	\centering 
	\includegraphics[width=1\textwidth, angle=0]{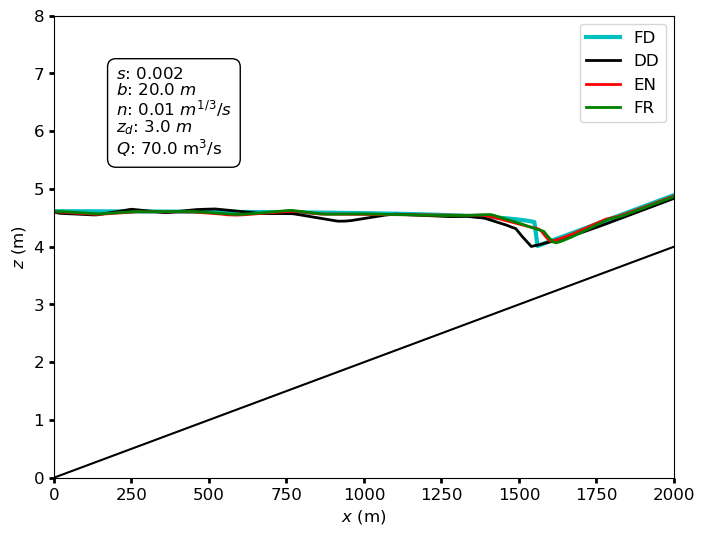}	
	\caption{Typical outcome of the comparison between the employed models, namely DD, EN, FR, and the reference solution, FD: only the case of the SP architecture is shown for the sake of clarity. The depicted profile, determined by the quantities in the box, contains a hydraulic jump.} 
	\label{SPex}
\end{figure}

In order to acquaint the reader with the physical problem at hand, Figure \ref{SPex} depicts a typical water level profile encompassing a hydraulic jump, as predicted by the SP, the latter having been trained with and without the inclusion of physical information. The reference solution, namely the profile resulting from the Finite Difference integration, is depicted as well. 
Incidentally, it is evident how the two physics-informed models outperform the purely DD one. A quantitative analysis is in order. 

In this section we show the results obtained from the application of the three different FFNN architectures, either in the purely DD or in the physics-informed training strategies. 

In the following, results yielded by the models in the initial configuration are presented and analyzed. The training has been carried out on the entire dataset. Then, the subsections contain results from several stress tests carried out by: 

i) reducing the size of the training dataset, while keeping the NN complexity fixed, thus mimicking a "small data regime"; 

ii) varying the NN complexity in a both data-scarce and data-rich scenario, therefore exploring both overfitting and underfitting conditions. 

iii) applying models for extrapolation, i.e. seeking predictions beyond the range of values covered during training, a scenario common to technical applications. 

In all the above tests, the effects of the physics-informed training strategy are evaluated. It is indeed widely reported in the literature \cite{eichelsdorfer2021physics, karniadakis2021physics, zhu2019physics} that the inclusion of physical information into ML models is beneficial especially in conditions of small data regimes.

We employed two key metrics to evaluate the performance in reconstructing water profiles, namely the Normalized Mean Absolute Error (NMAE) and the Normalized Nash-Sutcliffe Efficiency (NNSE). The NMAE, assuming the dam's height as a representative length scale for the flow depth, is defined as:
\begin{equation}
NMAE = \frac{\sum_{i=1}^{N} |h_i -\hat{h_i}| }{Nz_d} 
\label{MAE}
\end{equation}
The NNSE is formulated as:
\begin{equation}
NNSE = \frac{1}{(2-NSE)}
\label{NNSE}
\end{equation}
where
\begin{equation}
NSE = 1 - \frac{\sum_{i=1}^{N} (h_i - \hat{h}_i)^2}{\sum_{i=1}^{N} (h_i - \bar{h})^2}
\label{NSE}
\end{equation}
and $\bar{h}$ is the average depth of the profile. The rationale behind the scaling in Eq. \eqref{NNSE} is to bound the NSE value between $0$ and $1$, thus avoiding asymptotic tendency towards negative infinite values.

The NNSE is employed in addition to the NMAE due to its ability to amplify errors made close to the location of the hydraulic jump. Indeed, NNSE penalizes errors in areas where the flow depth is close to the mean value, and the hydraulic jump is bounded by the two conjugate depths which are the closest to the mean value within each profile.

A statistical analysis of the above two metrics shows a symmetrical shape of the cumulative frequencies distributions attained over the test set, thus allowing to employ their mean values as representative of the performance of each approach.

\subsection{Complete training dataset}

\begin{figure}
		\includegraphics[trim={0.5cm 0.5cm 0.5cm 0.5cm}, clip, width=\textwidth, angle=0]{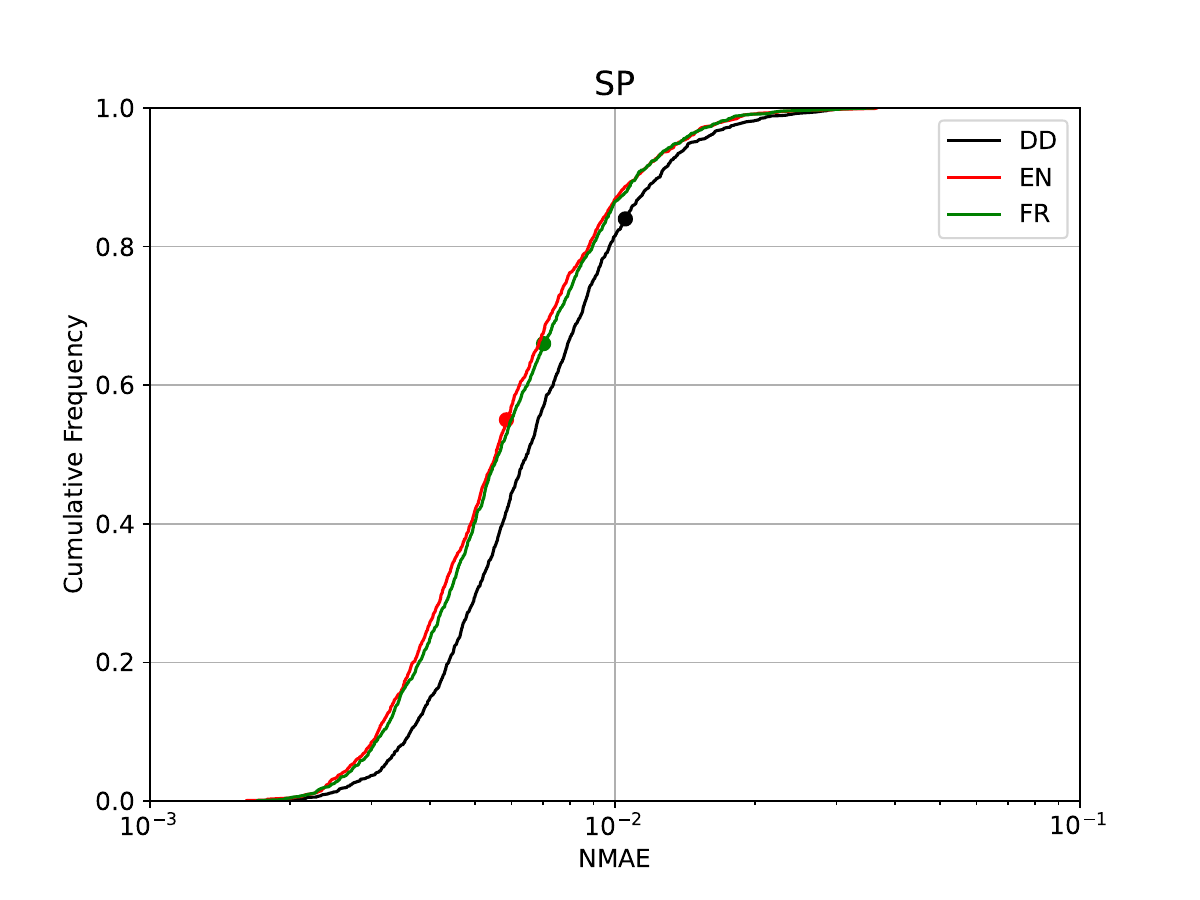}
		\includegraphics[trim={0.5cm 0.5cm 0.5cm 0.5cm}, clip,width=\textwidth, angle=0]{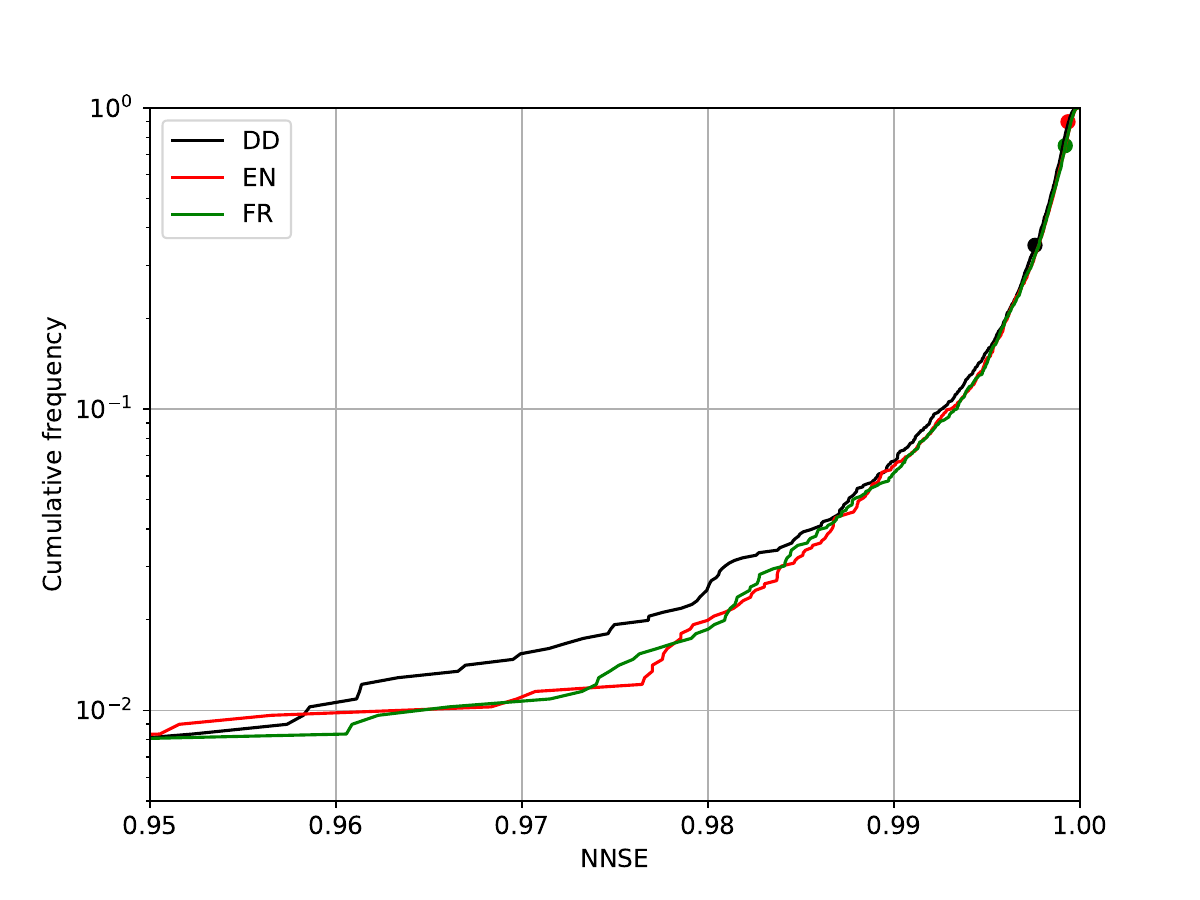}

	\caption{ Cumulative frequency distributions of NMAE and
NNSE values over the test set. Markers represent the positioning of the results for the example case shown in Figure \ref{SPex}} 
	\label{SP}
\end{figure}

The cumulative frequency distribution of NMAE and NNSE is employed to provide a comprehensive picture of the model performance over the whole test set.
In Figure \ref{SP} we depict the distributions for the SP architecture and the tree training strategies.
Due to their definitions in Eqs. \eqref{MAE} and \eqref{NNSE}, a better performing model features a higher cumulative frequency curve for NMAE and and a lower one for NNSE metrics. The values of the error metrics associated to the profiles shown in Figure \ref{SPex} are reported as markers in Figure \ref{SP} for the reader's convenience.

\begin{figure} 
	 \includegraphics[width=1\textwidth, angle=0]{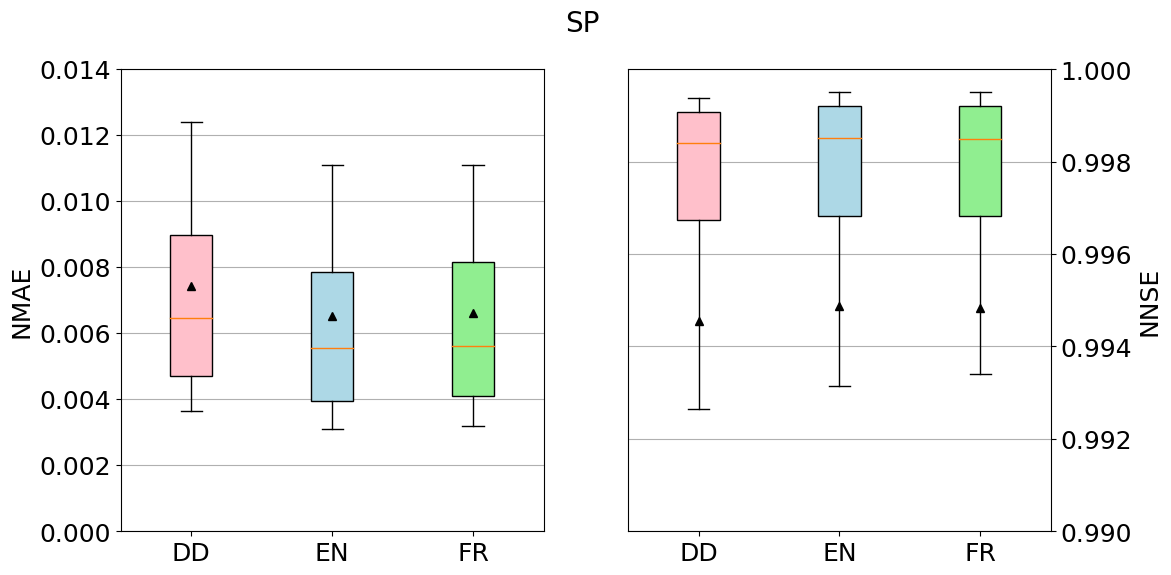}
      \includegraphics[width=1\textwidth, angle=0]{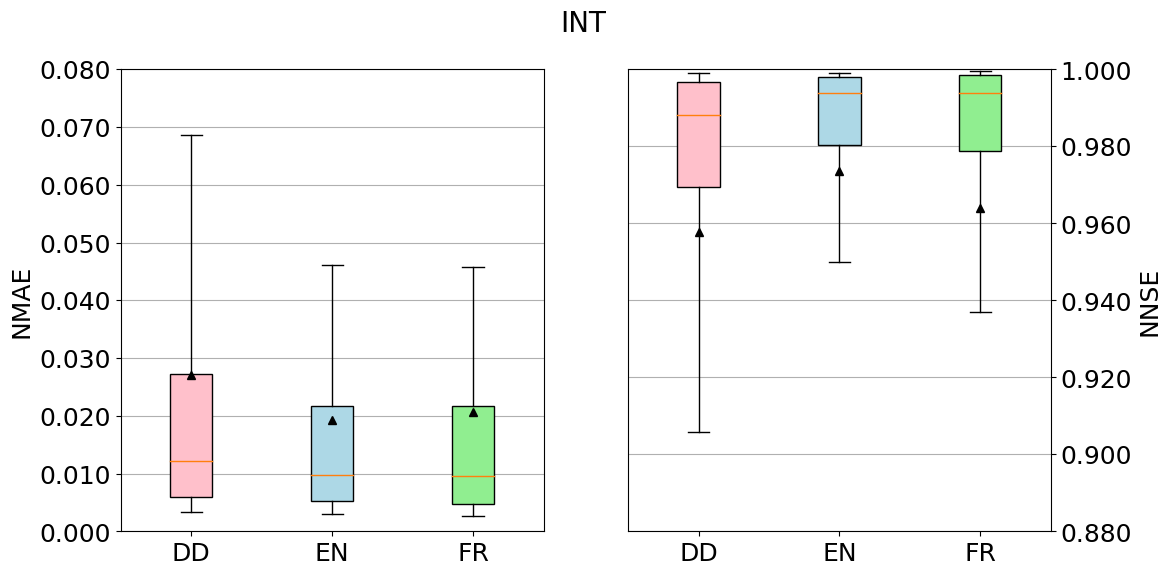}	
	\caption{Box plot of the NMAE and NNSE distributions for SP (upper panels) and INT (lower panels) architectures; boxes extend from the 25th to 75th percentile;  whiskers are placed at 10th and 90th percentiles; mean values and median values are respectively shown as triangles and orange segments. An overall improvement of predictive capabilities is detectable for both architecture employing physical training strategies. Results for VTS are shown in Figure \ref{VTSbox}. } 
	\label{INTbox}%
\end{figure} 

\begin{figure}
		\includegraphics[width=1.\textwidth, angle=0]{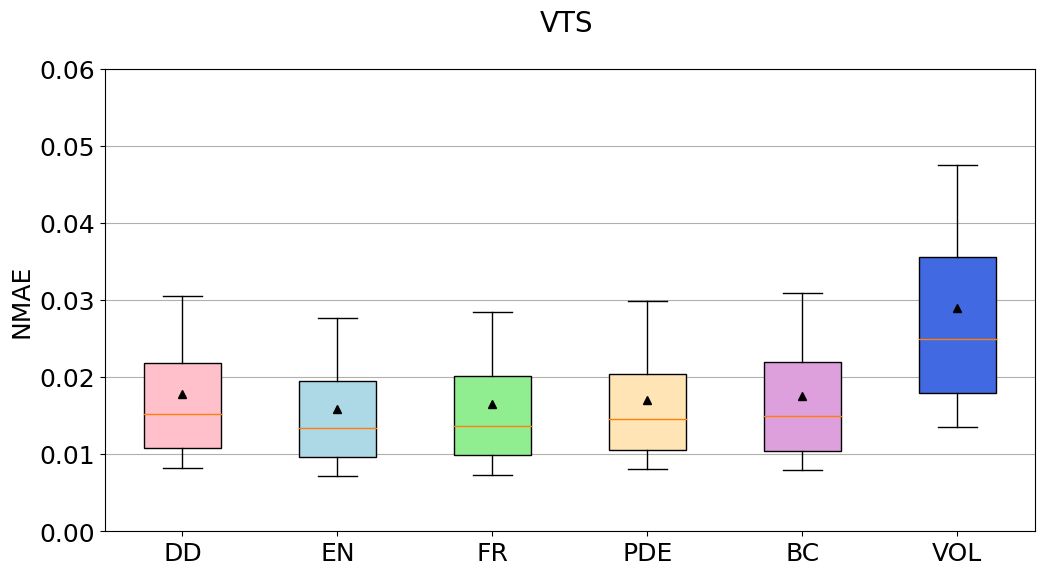}
		\includegraphics[width=1.\textwidth, angle=0]{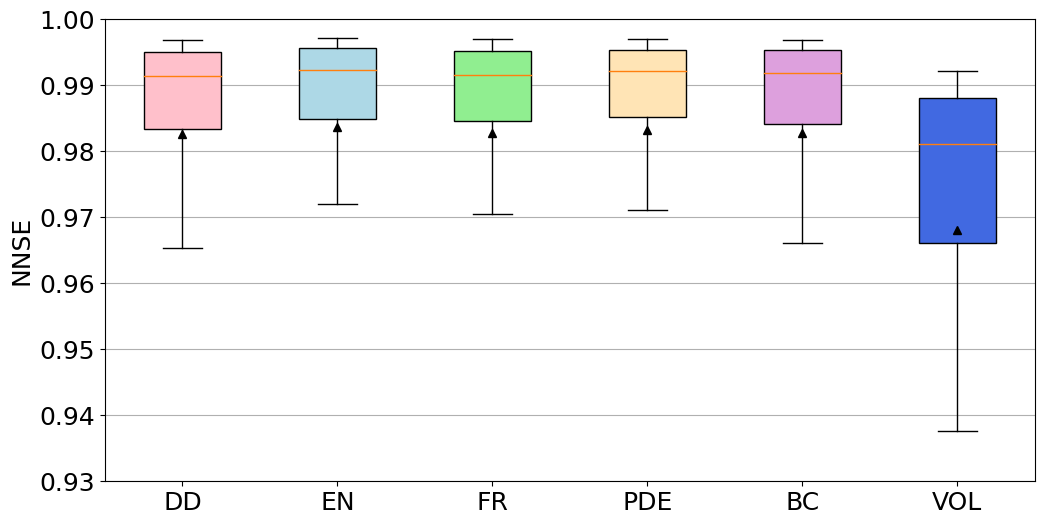}
 \caption{Box plot of the NMAE and NNSE distributions for the VTS architecture. Box plot features are the same as in Figure \ref{INTbox}. VTS also yields better predictions when trained with all types of the proposed physical constraints, but the volume (VOL) one.} 
\label{VTSbox}
\end{figure}

The overall picture suggests that the SP approach stands out with the best performance, while the integrator displays the highest variance in its results. The introduction of the physical information for all architectures enhances their performance, with minimal differences between EN and FR training strategies. In the context of VTS, all but the VOL training strategies induce a slight improvement in the model's performance (Figure \ref{VTSbox}). Constraining the volume of the water profile seems to consistently provide a deviant information content which detrimentally impacts the model's performance. Indeed, the same volume is shared by a large number of possible output profiles, so that the inclusion of the physical loss term results in the generation of additional local minima within the purely DD loss function. This example unveils that, despite apparently providing additional and physically sound information (i.e. the volume of water in the channel), the training phase may be diverted towards non physical solutions. This peculiar behavior could be shared with other physical constraints.

\subsection{Stress tests: unbalance between training dataset size and model complexity} 

We aim at training the above models so to induce either overfitting or underfitting. The former occurs when the model achieves a high fit to the training data while loses its ability to generalize to new unseen data. It can be fostered by either reducing the size of the training set or increasing the complexity of the model (i.e. the number of neurons within each layer). The reduction in the number of parameters is carried out while maintaining the three-layer structure and the same number of neurons among layers. The underfitting, conversely, manifests as a poor performance of the model caused by its lack of complexity with respect to the informational content of the training data. 

The data-scarce regime is obtained by excluding entire water profiles from the training dataset, since two of the three models (INT and VTS) require the stationing information to be retained in the reduced dataset.

\begin{figure}
	\includegraphics[width=1\textwidth, angle=0]{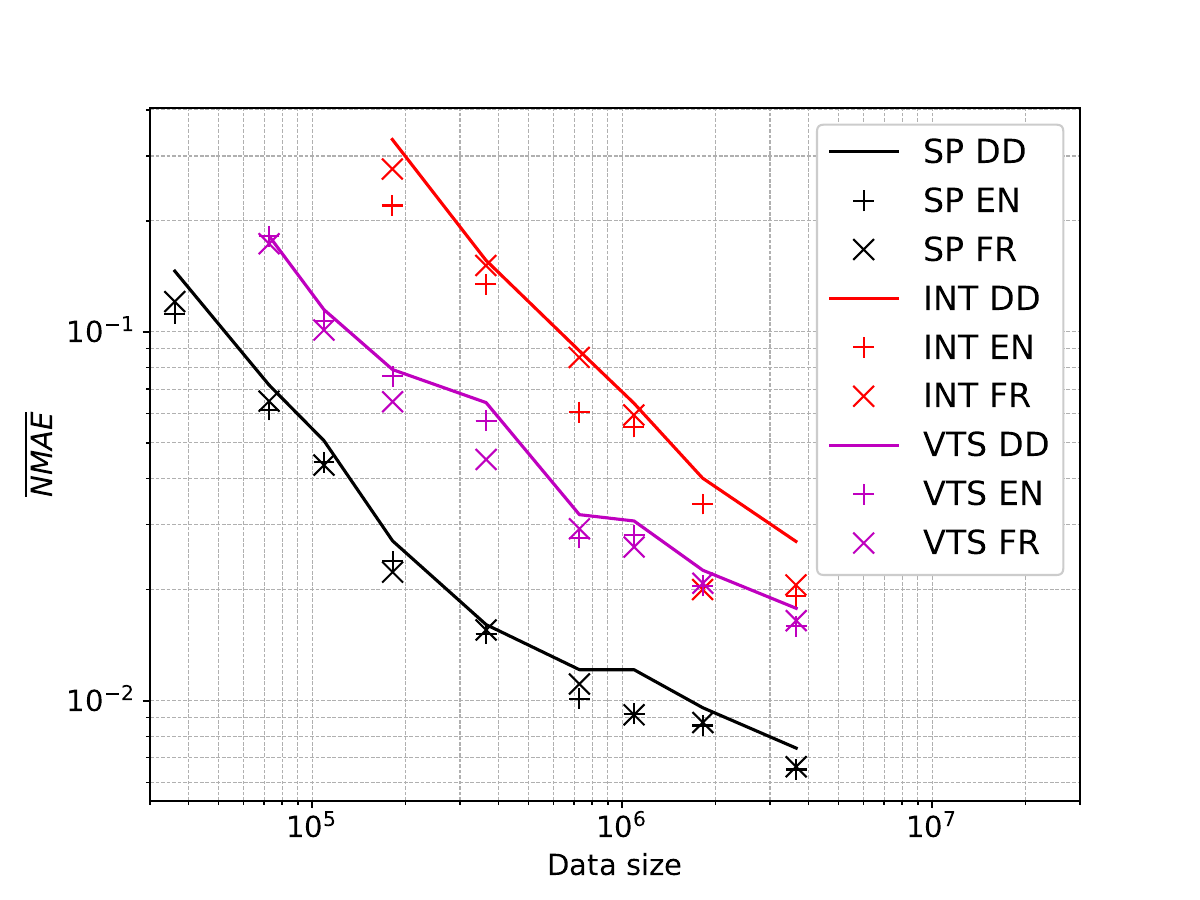}	
	\includegraphics[width=1\textwidth, angle=0]{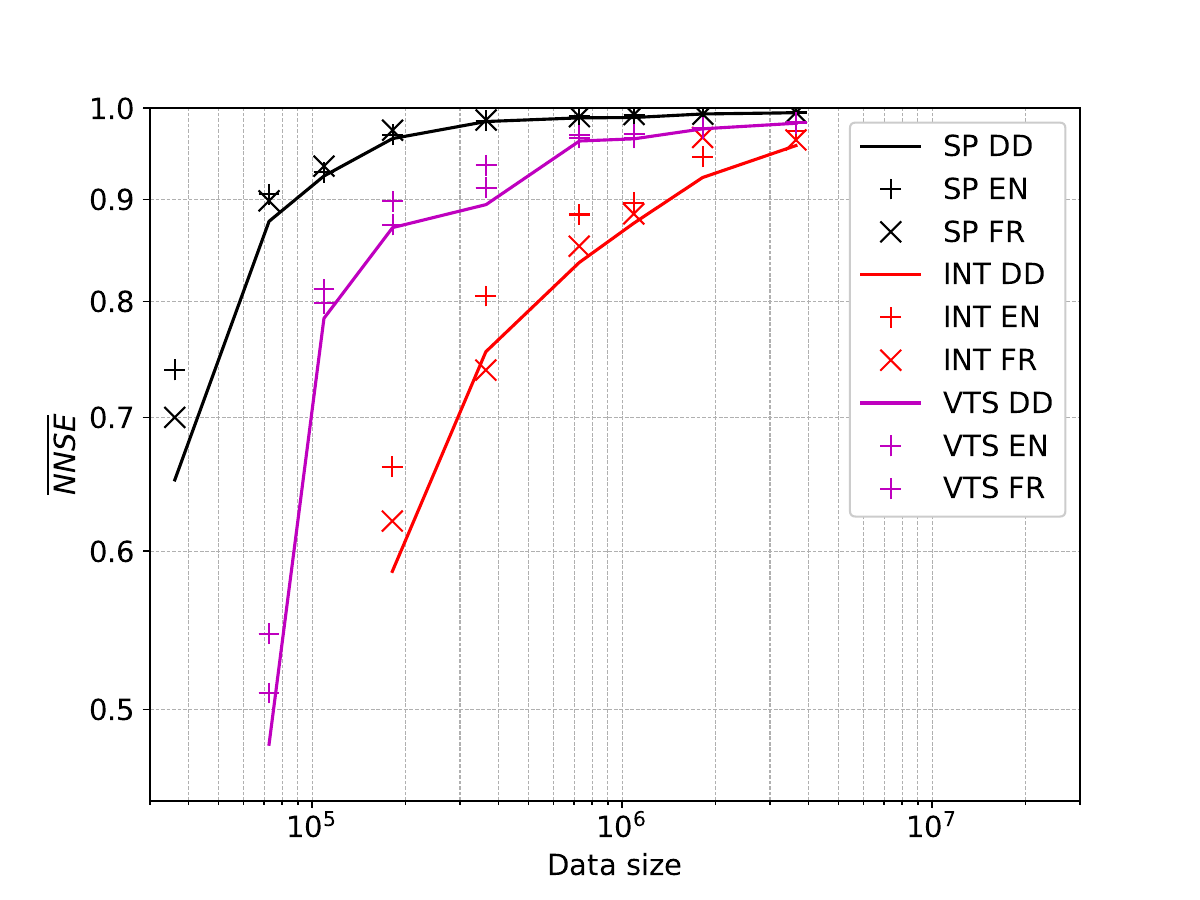}	
    \caption{Stress test: mean values of the distributions of NMAE and NNSE against the size of the training dataset. The physically based training strategies consistently mitigate the loss of accuracy of predictions when dataset size decreases. } 
    \label{small data stress}%
\end{figure}

In the following, in order to readily compare the performance of the models, only the average values of the cumulative frequency distributions of the two metrics are shown. Figure \ref{small data stress} shows that, while the performances progressively worsen by reducing the training set as expected, the physical training strategy improves the performance of all models and mitigates the overfitting, thus effectively enhancing generalization capabilities. The data-scarce scenario mimics a frequently occurring picture in the environmental hydraulics framework, when the response of scarcely gauged basins is sought.

Hereinafter, we vary the number of parameters in the models in both a data-scarce and data-rich regimes. The latter employs the whole available training dataset while the former only retains 5\%. 

\begin{figure}[H]
	\centering 
	\includegraphics[width=1\textwidth, angle=0]{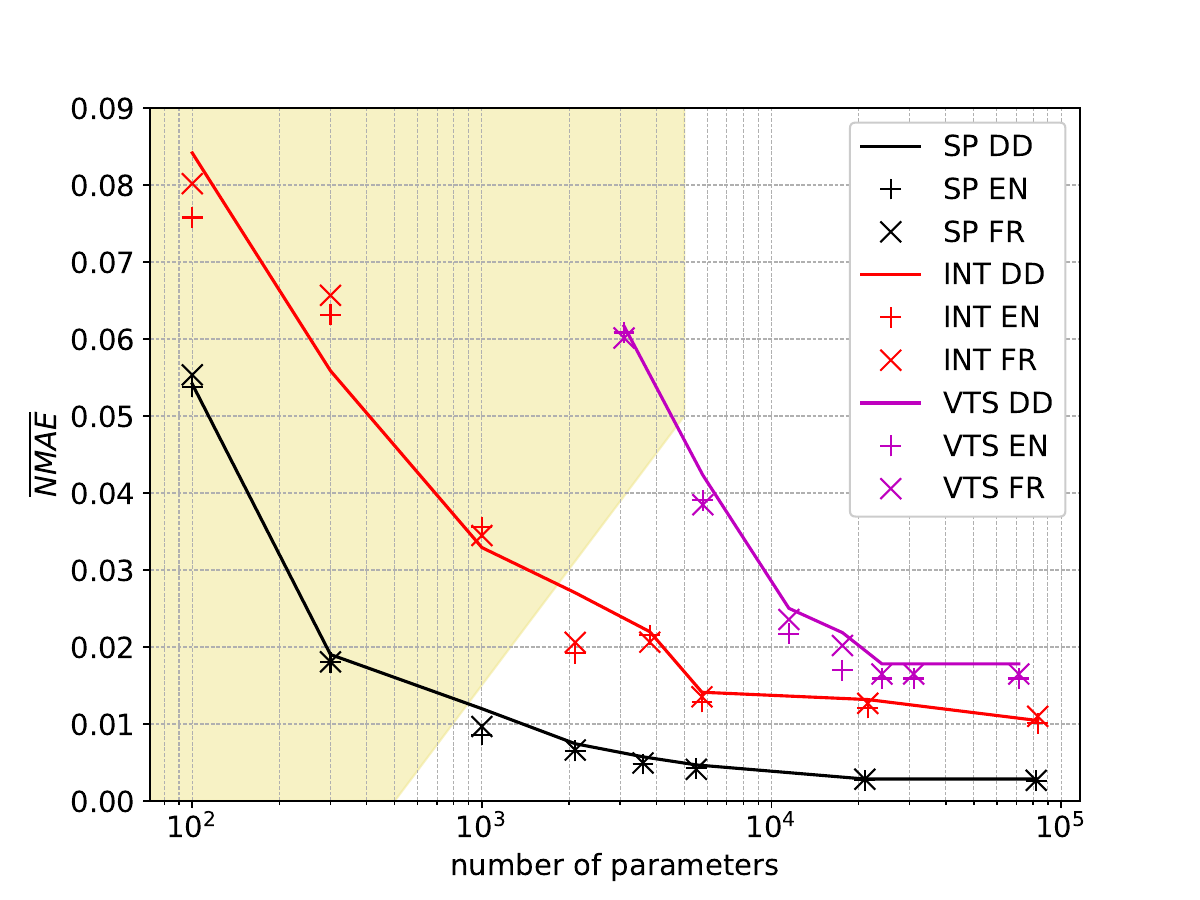}	
    \includegraphics[width=1\textwidth, angle=0]{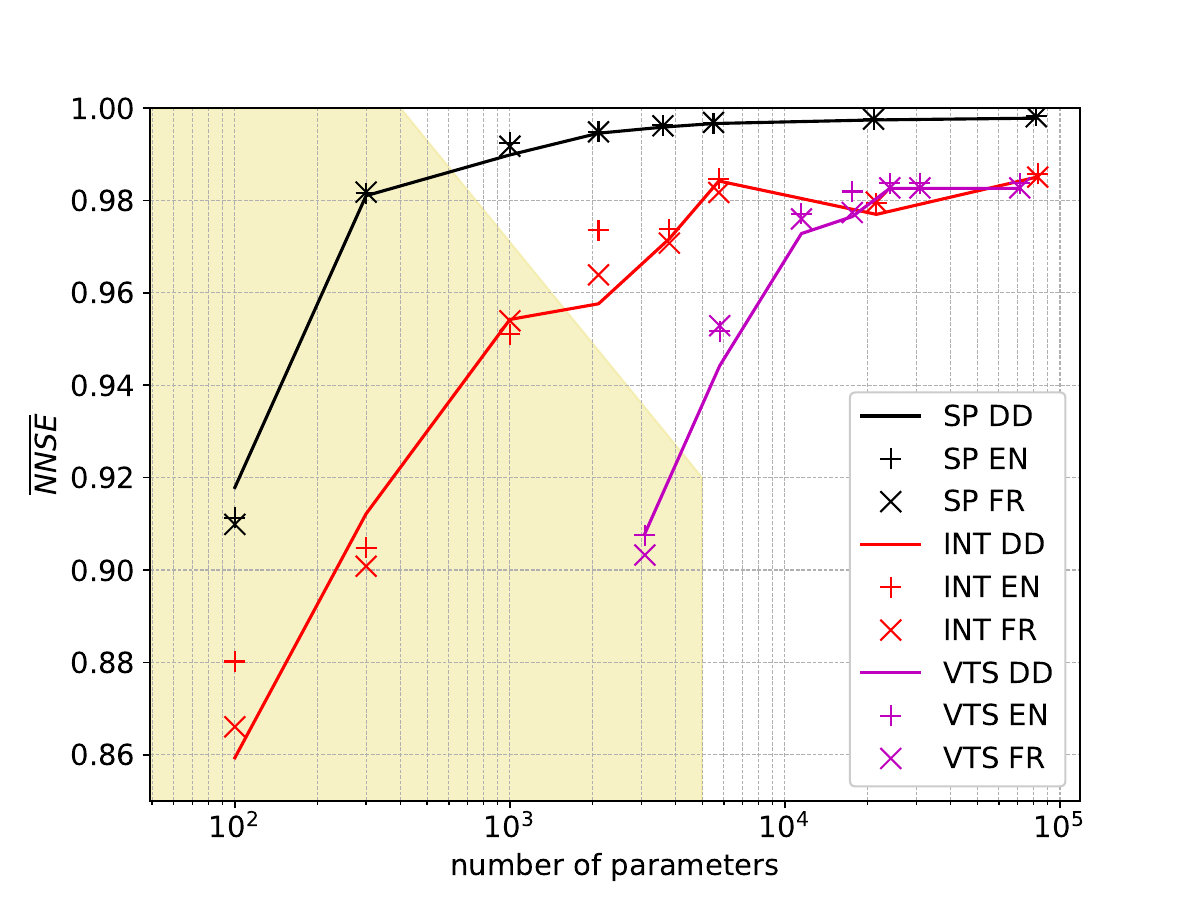}	
	\caption{Stress test: mean values of the distributions of NMAE and NNSE against  models' complexity (full training dataset). In the yellow-shaded region (underfitting regime), physical constraints do not consistently lead to improvements.} 
	\label{NMAE NP}%
\end{figure}

In Figure \ref{NMAE NP}, an expected trend emerges as the number of parameters decreases: performance consistently deteriorates across all three architectures. This discernible pattern clearly unveils the occurrence of underfitting. The lower complexity architectures now struggle to capture the variability spanned by the complete dataset. As a consequence, the models are insufficiently expressive, resulting in suboptimal performance metrics.
In such underfitting regime, it is noteworthy that the introduction of a physical training strategy does not yield a consistent improvement in generalization results. This behavior is the consequence of the requirement for the low-complexity model, which is already struggling to discern patterns in the data, to fulfill additional constraints, such as the physical ones.

Instead, with reference to Figure \ref{NMAE NP}, it is evident that when increasing model complexity up to the occurrence of overfitting (on the right of the elbow of the curve), the physics-informed models consistently yield better results than the purely DD ones.

\begin{figure}[H]
	\centering 
	\includegraphics[width=1\textwidth, angle=0]{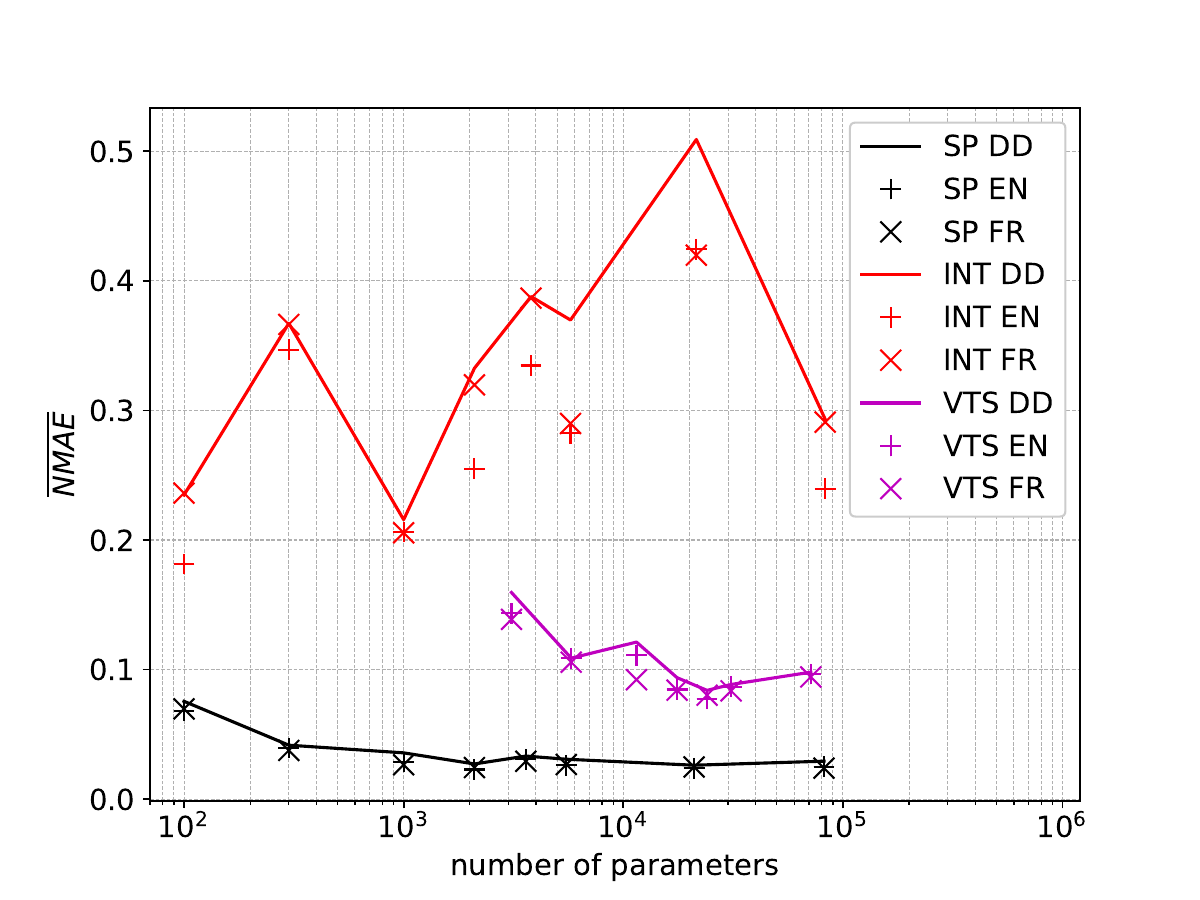}
     \includegraphics[width=1\textwidth, angle=0]{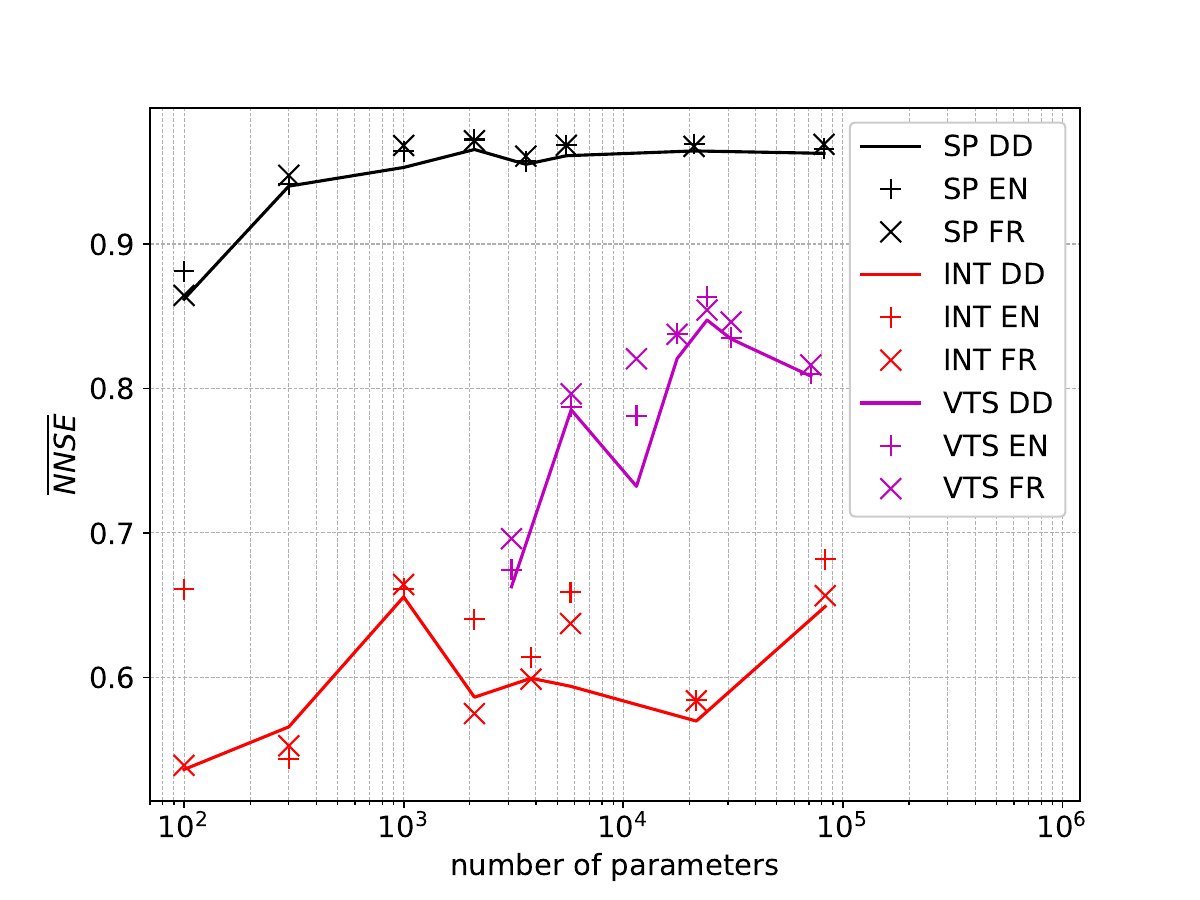}
 	\caption{Stress test: mean values of the distributions of NMAE and NNSE against  models' complexity  (5\% of training data). In data scarce scenario, the beneficial effect of adding the physical information emerges.} 
	\label{NMAE NP 5}%
\end{figure}

In the context of a small data regime, however, different behaviors emerge. Figure \ref{NMAE NP 5} illustrates that, in situations with limited available data, model performance shows loose dependence on the number of parameters and, in some cases, it even deteriorates as the complexity of the models increases. Also in this clear scenario of overfitting, models enhanced with the physical training strategy consistently ensure an improvement in performance. The inclusion of the physical loss term acts as a regularization mechanism, improving generalization. 
Moreover, the physical term involves both input and output data, akin to a form of data augmentation, providing the model with further insights into the nature of the system to be interpreted. 

\subsection{Extrapolation}
\begin{figure}
	\centering 
	\includegraphics[trim={1cm 1cm 0.5cm 0.5cm}, clip, width=1.\textwidth, angle=0]{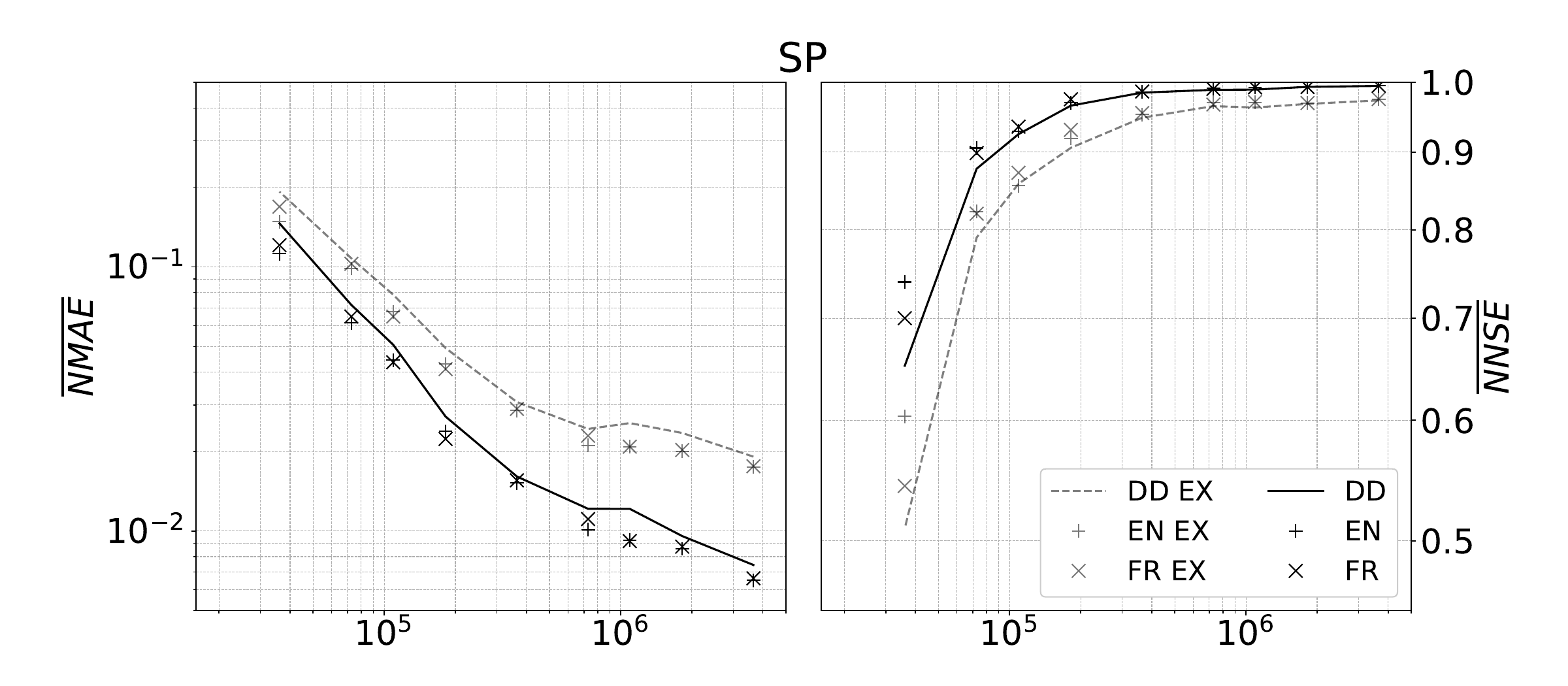}
    \includegraphics[trim={1cm 1cm 0.5cm 0.5cm}, clip, width=1.\textwidth, angle=0]{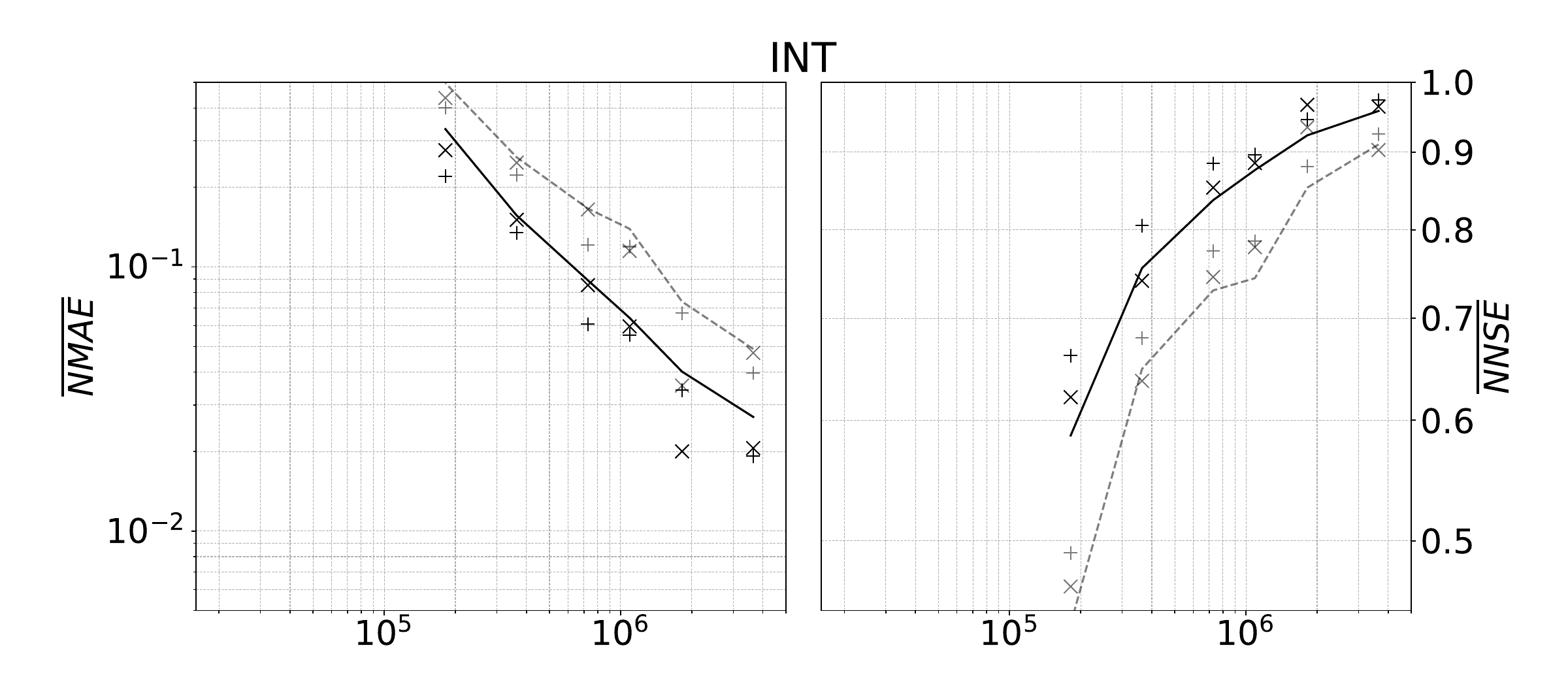}
    \includegraphics[trim={1cm 0cm 0.5cm 0.5cm}, clip, width=1.\textwidth, angle=0]{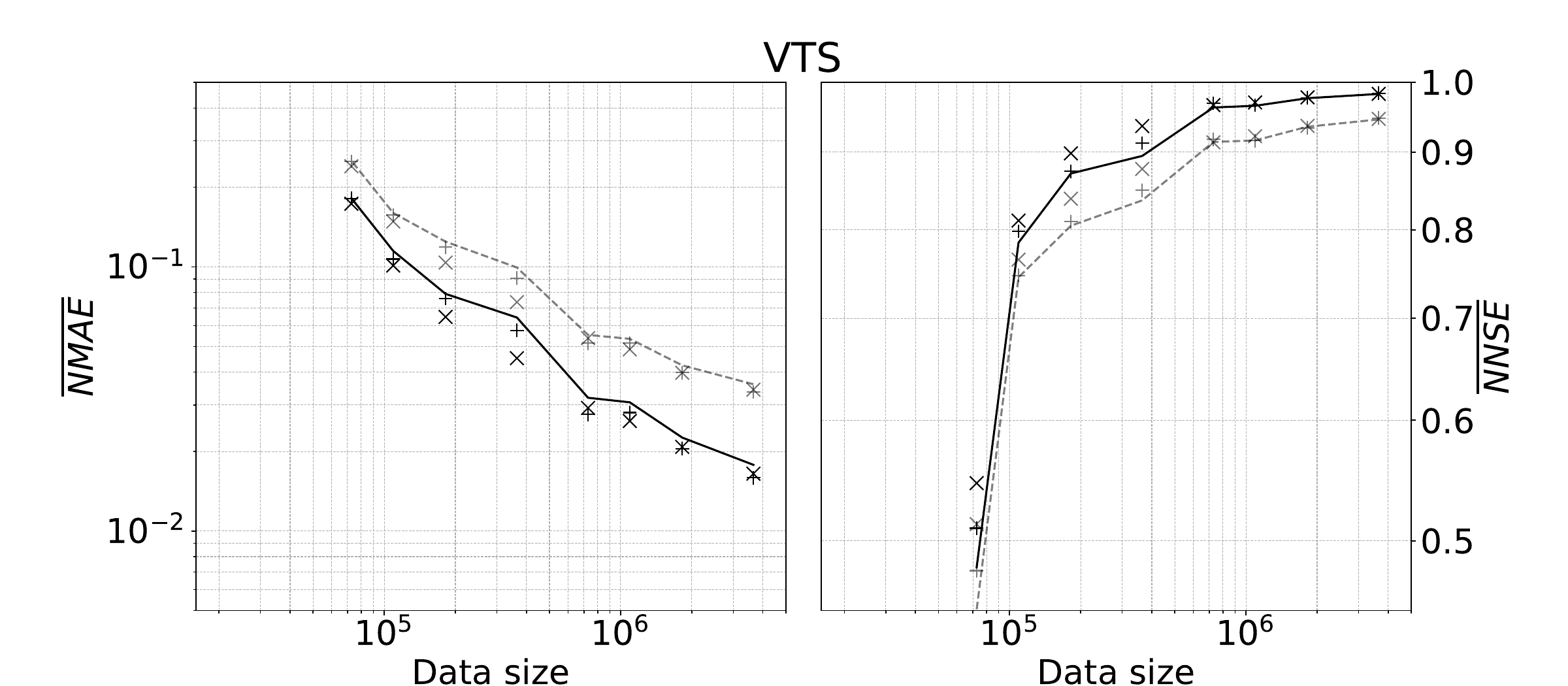}
	\caption{Stress test: assessment of models' accuracy in predicting scenarios out of the range of the training dataset (extrapolation, EX series, depicted with light traces and markers); each panels show the same stress test results for the models on scenarios encompassed by the trained dataset (same data shown in \ref{small data stress}, depicted with bold traces and markers). Though performance is lower for extrapolated scenarios, the beneficial effect of physical information is still evident.} 
	\label{nmae_ext}%
\end{figure}

Contrary to conventional practices in machine learning frameworks, we also explore the extrapolation abilities of the models. New profiles were generated by considering at least one of the values of the five inputs $s, b, n, z_d, Q$ extending by 10\% above or below the range covered by the training set. Out of these profiles, 1558 were randomly selected (same size of test set), obtaining a modified test set (referred to as \textit{EXT}).
The performance curves of the models on the \textit{EXT} set when decreasing the size of the training set (as previously seen for the small data regime stress test), are illustrated in Figure \ref{nmae_ext}. These curves are compared with the results obtained using the standard test set (interpolation). Beside the expected drop in overall performance compared to the interpolation cases, the results confirm that models trained with physical information exhibit greater generalization capabilities. These results are of crucial importance for the perspective application to flood mapping. Indeed, in such field predictions are often sought not only for scenarios falling within the quantitative range of available observations, though previously unexplored, but also for cases featuring values of the observed quantities falling out of the range of the recorded series.

\section{Conclusions}
\label{Conclusions}
In this work, we propose the incorporation of \textit{soft} physical information in the training strategy of \textit{neural operators} to improve their generalization capabilities in the context of environmental hydraulics.

\textit{Neural operators} are designed to mimic the relationships between input and output functions across an entire class of physical phenomena, as opposed to \textit{neural solvers}, which are aimed at finding the solution to a specific realization. Indeed, the latter approach may not be useful in environmental hydraulics, where a large number of scenarios need to be explored.

Moreover, the proposed \textit{soft} physical information does not need to be formulated in terms of the governing PDEs, providing a considerable advantage in presence of large epistemic uncertainty, as in river hydraulics.

We tested the effect of soft physical information on a controllable and yet highly informative test case commonly used in hydraulics, namely a 1D steady state, mixed-regime, water profile in a rectangular channel. This physical system often develops a discontinuity, represented by the hydraulic jump, whose solution challenges the employed models.

Three NN architectures were trained on synthetic data alone and in conjunction with soft physical information.
The latter provide better predictive skills, particularly in the presence of overfitting. Underfitting does not seem to be mitigated by the added physical information. 
Moreover, in order to replicate a common practice in the environmental modeling, the NNs predictions are evaluated also for scenarios falling beyond the range of the training data (extrapolation). Even in this case, the physical information allows for a measurable gain in performance. 
These results are of great relevance to the possible applications of NNs to flood mapping, where cases featuring values of the observed quantities falling out of the range of the recorded series need to be predicted.
Finally, our analysis unveils the possible occurrence of a detrimental effect of an apparently informative physical constraint. In this sense, our work calls for the development of a systematic framework to measure the informational content of physical constraints.

\section*{Acknowledgments}
We sincerely thank Dr. Andrea Gemma for his valuable support in setting up the initial version of the code.

\section*{Funding}
The G. G. doctoral scholarship is granted under the National Recovery and Resilience Plan (PNRR, NextGenerationEU).

J.-M. T. thanks the FRQNT “Fonds de recherche du Québec – Nature et technologies (FRQNT)" for financial support (Research Scholarship No. 314328).

\section*{Declaration of interests}
None

\appendix
\section{Neural Networks: Architectures and Training}
\label{FFNN RNN}
Neural Networks were inspired by the architecture of nerve cells. A single element of the network, called neuron, receives a series of $I$ inputs $a_i$, calculates a weighted sum of these inputs (which may include a bias value $c$), and then applies an activation function $\sigma$ to produce the neuron's output $z$.
\begin{equation}
z = \sigma\left(\sum_{i=1}^{I} w_i a_i + c \right)
\label{neuron}
\end{equation}
Among the most commonly used activation functions are the Sigmoid, the Hyperbolic Tangent, the Rectified Linear Unit (ReLU), and the Leaky ReLU \cite{apicella2021survey} function. The choice of the activation function affects the model outcome.

NNs can be categorised into Feed-Forward Neural Networks (FFNN), where the signal propagates only from input to output, and Recurrent Neural Networks (RNN), where the output of a neuron is fed back as its input. Both are composed of an input layer, a series of $H$ hidden layers, and an output layer. All networks employed in this work are FFNNs. 

The approximate function $F$ represented by a FFNN can then be seen as the composition of several functions $f$, one for each hidden layer:
\begin{equation}
F=f_H \circ f_{H-1} \circ ... \circ f_{1} 
\label{composition}
\end{equation}

A FFNN with more than one layer can theoretically represent any continuous, or even discontinuous function, to arbitrary precision, given a sufficient number of neurons. 

The weights optimization phase (training) of NNs exploits a gradient descent optimizer, such as the Adam algorithm \cite{kingma2014adam}, to iteratively find the minimum of the loss function. The learning rate determines the step size taken during each iteration of the optimization process and plays a crucial role to ensure convergence. The gradient descent algorithm needs to evaluate the gradients $\partial{\mathcal{L}} / \partial{w}$ of the loss function with respect to every weight $w$ in the network. Such computation is carried out by means of the Backpropagation algorithm \cite{rumelhart1986learning}, which efficiently calculates these gradients by propagating the error backward through the network.

\section{Dataset generation}
\label{dataset generation}

The dataset was generated by solving the specific energy equation, expressed as \cite{cengel2013ebook}:
\begin{equation}
\frac{dE}{dx} = i - J
\label{energy equation}
\end{equation}
Here, 
$E$ represents the specific energy relative to the channel bottom, which, for a rectangular channel, can be calculated as:
\begin{equation}
E = h + \frac{Q^2}{2gb^2h^2}
\label{specific energy}
\end{equation}
$g$ is the gravity,
$b$ is the channel width, 
$h$ is the flow depth;
$J$ is the energy grade slope, calculated using the Chezy relation:
\begin{equation}
J = \frac{n^2 Q^2}{\Omega^2R^{4/3}}
\label{Chezy}
\end{equation}
$n$ is the Manning coefficient,
$\Omega = bh $ is the cross-sectional flow area and $R=\Omega/(b+2h)$ is the hydraulic radius.
The downstream boundary condition is represented by the broad-crested weir equation:
\begin{equation}
h = z_d + \left( \frac{3 \sqrt{3} Q}{2 \sqrt{2 g} b} \right)^{2/3}
\label{weir}
\end{equation}
where $z_d$ is the weir height.

The possible transition from supercritical ($Fr > 1$) to subcritical ($Fr < 1$) flow occurs through a hydraulic jump, where:
\begin{equation}
Fr = \frac{Q}{\Omega \sqrt{g h}}
\label{Froude}
\end{equation}

Examining the conservation of momentum within the fluid volume encompassing the hydraulic jump, two cross-sections denoted as 1 (upstream, supercritical flow) and 2 (downstream, subcritical flow) can be defined \cite{cengel2013ebook}. The conservation of momentum can be expressed as:

\begin{equation}
\Pi_{p1} + M_1 = \Pi_{p2} + M_2
\label{momentum}
\end{equation}

Here, $\Pi_{p1}$ and $\Pi_{p2}$ represent the upstream and downstream values of the hydrostatic force, while $M_1$ and $M_2$ denote the upstream and downstream values of the momentum flux, respectively. 
For a rectangular prismatic channel, equation \eqref{momentum} becomes:
\begin{equation}
\frac{y_1^2}{2}+\frac{Q^2}{g b^2 y_1} = \frac{y_2^2}{2}+\frac{Q^2}{g b^2 y_2}
\label{momentum rectangular}
\end{equation}
which can be arranged in the following expression, establishing a mathematical relation between $y_1$ and $y_2$, that are called conjugate depths:
\begin{equation}
y_1 = \frac{y_2}{2}(-1+\sqrt{1+8F_2^2})
\label{conjugate depths}
\end{equation}
$F_2$ is the Froude number (equation \eqref{Froude}) at the cross-section 2. 

The water profile is obtained by solving the differential equation \eqref{energy equation} using a first order FD scheme, with a constant spatial discretization $\Delta x$. An upstream marching algorithm is employed, encompassing the solution of the hydraulic jump if necessary.

\bibliographystyle{elsarticle-harv} 
\bibliography{main}

\begin{thebibliography}{41}
\expandafter\ifx\csname natexlab\endcsname\relax\def\natexlab#1{#1}\fi
\providecommand{\url}[1]{\texttt{#1}}
\providecommand{\href}[2]{#2}
\providecommand{\path}[1]{#1}
\providecommand{\DOIprefix}{doi:}
\providecommand{\ArXivprefix}{arXiv:}
\providecommand{\URLprefix}{URL: }
\providecommand{\Pubmedprefix}{pmid:}
\providecommand{\doi}[1]{\href{http://dx.doi.org/#1}{\path{#1}}}
\providecommand{\Pubmed}[1]{\href{pmid:#1}{\path{#1}}}
\providecommand{\bibinfo}[2]{#2}
\ifx\xfnm\relax \def\xfnm[#1]{\unskip,\space#1}\fi
%Type = Misc
\bibitem[{Abadi et~al.(2015)Abadi, Agarwal, Barham, Brevdo, Chen, Citro, Corrado, Davis, Dean, Devin, Ghemawat, Goodfellow, Harp, Irving, Isard, Jia, Jozefowicz, Kaiser, Kudlur, Levenberg, Man\'{e}, Monga, Moore, Murray, Olah, Schuster, Shlens, Steiner, Sutskever, Talwar, Tucker, Vanhoucke, Vasudevan, Vi\'{e}gas, Vinyals, Warden, Wattenberg, Wicke, Yu and Zheng}]{tensorflow2015-whitepaper}
\bibinfo{author}{Abadi, M.}, \bibinfo{author}{Agarwal, A.}, \bibinfo{author}{Barham, P.}, \bibinfo{author}{Brevdo, E.}, \bibinfo{author}{Chen, Z.}, \bibinfo{author}{Citro, C.}, \bibinfo{author}{Corrado, G.S.}, \bibinfo{author}{Davis, A.}, \bibinfo{author}{Dean, J.}, \bibinfo{author}{Devin, M.}, \bibinfo{author}{Ghemawat, S.}, \bibinfo{author}{Goodfellow, I.}, \bibinfo{author}{Harp, A.}, \bibinfo{author}{Irving, G.}, \bibinfo{author}{Isard, M.}, \bibinfo{author}{Jia, Y.}, \bibinfo{author}{Jozefowicz, R.}, \bibinfo{author}{Kaiser, L.}, \bibinfo{author}{Kudlur, M.}, \bibinfo{author}{Levenberg, J.}, \bibinfo{author}{Man\'{e}, D.}, \bibinfo{author}{Monga, R.}, \bibinfo{author}{Moore, S.}, \bibinfo{author}{Murray, D.}, \bibinfo{author}{Olah, C.}, \bibinfo{author}{Schuster, M.}, \bibinfo{author}{Shlens, J.}, \bibinfo{author}{Steiner, B.}, \bibinfo{author}{Sutskever, I.}, \bibinfo{author}{Talwar, K.}, \bibinfo{author}{Tucker, P.}, \bibinfo{author}{Vanhoucke, V.}, \bibinfo{author}{Vasudevan, V.},
  \bibinfo{author}{Vi\'{e}gas, F.}, \bibinfo{author}{Vinyals, O.}, \bibinfo{author}{Warden, P.}, \bibinfo{author}{Wattenberg, M.}, \bibinfo{author}{Wicke, M.}, \bibinfo{author}{Yu, Y.}, \bibinfo{author}{Zheng, X.}, \bibinfo{year}{2015}.
\newblock \bibinfo{title}{{TensorFlow}: Large-scale machine learning on heterogeneous systems}.
\newblock \URLprefix \url{https://www.tensorflow.org/}. \bibinfo{note}{software available from tensorflow.org}.
%Type = Article
\bibitem[{Ali et~al.(2014)Ali, Faraj, Koya, Ali and Faraj}]{ali2014data}
\bibinfo{author}{Ali, P.J.M.}, \bibinfo{author}{Faraj, R.H.}, \bibinfo{author}{Koya, E.}, \bibinfo{author}{Ali, P.J.M.}, \bibinfo{author}{Faraj, R.H.}, \bibinfo{year}{2014}.
\newblock \bibinfo{title}{Data normalization and standardization: a technical report}.
\newblock \bibinfo{journal}{Mach Learn Tech Rep} \bibinfo{volume}{1}, \bibinfo{pages}{1--6}.
%Type = Article
\bibitem[{Apicella et~al.(2021)Apicella, Donnarumma, Isgr{\`o} and Prevete}]{apicella2021survey}
\bibinfo{author}{Apicella, A.}, \bibinfo{author}{Donnarumma, F.}, \bibinfo{author}{Isgr{\`o}, F.}, \bibinfo{author}{Prevete, R.}, \bibinfo{year}{2021}.
\newblock \bibinfo{title}{A survey on modern trainable activation functions}.
\newblock \bibinfo{journal}{Neural Networks} \bibinfo{volume}{138}, \bibinfo{pages}{14--32}.
%Type = Article
\bibitem[{Baydin et~al.(2018)Baydin, Pearlmutter, Radul and Siskind}]{baydin2018automatic}
\bibinfo{author}{Baydin, A.G.}, \bibinfo{author}{Pearlmutter, B.A.}, \bibinfo{author}{Radul, A.A.}, \bibinfo{author}{Siskind, J.M.}, \bibinfo{year}{2018}.
\newblock \bibinfo{title}{Automatic differentiation in machine learning: a survey}.
\newblock \bibinfo{journal}{Journal of Marchine Learning Research} \bibinfo{volume}{18}, \bibinfo{pages}{1--43}.
%Type = Article
\bibitem[{Bentivoglio et~al.(2022)Bentivoglio, Isufi, Jonkman and Taormina}]{bentivoglio2022deep}
\bibinfo{author}{Bentivoglio, R.}, \bibinfo{author}{Isufi, E.}, \bibinfo{author}{Jonkman, S.N.}, \bibinfo{author}{Taormina, R.}, \bibinfo{year}{2022}.
\newblock \bibinfo{title}{Deep learning methods for flood mapping: a review of existing applications and future research directions}.
\newblock \bibinfo{journal}{Hydrology and Earth System Sciences} \bibinfo{volume}{26}, \bibinfo{pages}{4345--4378}.
%Type = Article
\bibitem[{Berkhahn et~al.(2019)Berkhahn, Fuchs and Neuweiler}]{berkhahn2019ensemble}
\bibinfo{author}{Berkhahn, S.}, \bibinfo{author}{Fuchs, L.}, \bibinfo{author}{Neuweiler, I.}, \bibinfo{year}{2019}.
\newblock \bibinfo{title}{An ensemble neural network model for real-time prediction of urban floods}.
\newblock \bibinfo{journal}{Journal of Hydrology} \bibinfo{volume}{575}, \bibinfo{pages}{743--754}.
%Type = Book
\bibitem[{Bl{\"o}schl(2013)}]{bloschl2013runoff}
\bibinfo{author}{Bl{\"o}schl, G.}, \bibinfo{year}{2013}.
\newblock \bibinfo{title}{Runoff prediction in ungauged basins: synthesis across processes, places and scales}.
\newblock \bibinfo{publisher}{Cambridge University Press}.
%Type = Article
\bibitem[{Cai et~al.(2021)Cai, Mao, Wang, Yin and Karniadakis}]{cai2021physics}
\bibinfo{author}{Cai, S.}, \bibinfo{author}{Mao, Z.}, \bibinfo{author}{Wang, Z.}, \bibinfo{author}{Yin, M.}, \bibinfo{author}{Karniadakis, G.E.}, \bibinfo{year}{2021}.
\newblock \bibinfo{title}{Physics-informed neural networks (pinns) for fluid mechanics: A review}.
\newblock \bibinfo{journal}{Acta Mechanica Sinica} \bibinfo{volume}{37}, \bibinfo{pages}{1727--1738}.
%Type = Article
\bibitem[{Cedillo et~al.(2022)Cedillo, N{\'u}{\~n}ez, S{\'a}nchez-Cordero, Timbe, Samaniego and Alvarado}]{cedillo2022physics}
\bibinfo{author}{Cedillo, S.}, \bibinfo{author}{N{\'u}{\~n}ez, A.G.}, \bibinfo{author}{S{\'a}nchez-Cordero, E.}, \bibinfo{author}{Timbe, L.}, \bibinfo{author}{Samaniego, E.}, \bibinfo{author}{Alvarado, A.}, \bibinfo{year}{2022}.
\newblock \bibinfo{title}{Physics-informed neural network water surface predictability for 1d steady-state open channel cases with different flow types and complex bed profile shapes}.
\newblock \bibinfo{journal}{Advanced Modeling and Simulation in Engineering Sciences} \bibinfo{volume}{9}, \bibinfo{pages}{1--23}.
%Type = Book
\bibitem[{Cengel and Cimbala(2013)}]{cengel2013ebook}
\bibinfo{author}{Cengel, Y.}, \bibinfo{author}{Cimbala, J.}, \bibinfo{year}{2013}.
\newblock \bibinfo{title}{Ebook: Fluid mechanics fundamentals and applications (si units)}.
\newblock \bibinfo{publisher}{McGraw Hill}.
%Type = Article
\bibitem[{Chiu et~al.()Chiu, Wong, Ooi, Dao and Ong}]{chiumarrying}
\bibinfo{author}{Chiu, P.H.}, \bibinfo{author}{Wong, J.C.}, \bibinfo{author}{Ooi, C.C.}, \bibinfo{author}{Dao, M.H.}, \bibinfo{author}{Ong, Y.S.}, .
\newblock \bibinfo{title}{Marrying the benefits of automatic and numerical differentiation in physics-informed neural network} .
%Type = Misc
\bibitem[{Chollet et~al.(2015)}]{chollet2015keras}
\bibinfo{author}{Chollet, F.}, et~al., \bibinfo{year}{2015}.
\newblock \bibinfo{title}{Keras}.
\newblock \bibinfo{howpublished}{\url{https://keras.io}}.
%Type = Inproceedings
\bibitem[{Cole et~al.(2006)Cole, Moore, Bell and Jones}]{cole2006issues}
\bibinfo{author}{Cole, S.}, \bibinfo{author}{Moore, R.}, \bibinfo{author}{Bell, V.}, \bibinfo{author}{Jones, D.}, \bibinfo{year}{2006}.
\newblock \bibinfo{title}{Issues in flood forecasting: ungauged basins, extreme floods and uncertainty}, in: \bibinfo{booktitle}{Frontiers in Flood Forecasting, 8th Kovacs Colloquium, UNESCO, Paris}, pp. \bibinfo{pages}{103--122}.
%Type = Article
\bibitem[{Cuomo et~al.(2022)Cuomo, Di~Cola, Giampaolo, Rozza, Raissi and Piccialli}]{cuomo2022scientific}
\bibinfo{author}{Cuomo, S.}, \bibinfo{author}{Di~Cola, V.S.}, \bibinfo{author}{Giampaolo, F.}, \bibinfo{author}{Rozza, G.}, \bibinfo{author}{Raissi, M.}, \bibinfo{author}{Piccialli, F.}, \bibinfo{year}{2022}.
\newblock \bibinfo{title}{Scientific machine learning through physics--informed neural networks: Where we are and what’s next}.
\newblock \bibinfo{journal}{Journal of Scientific Computing} \bibinfo{volume}{92}, \bibinfo{pages}{88}.
%Type = Article
\bibitem[{Dasgupta et~al.(2023)Dasgupta, Das, Banerjee and Mazumdar}]{dasgupta2023revisit}
\bibinfo{author}{Dasgupta, R.}, \bibinfo{author}{Das, S.}, \bibinfo{author}{Banerjee, G.}, \bibinfo{author}{Mazumdar, A.}, \bibinfo{year}{2023}.
\newblock \bibinfo{title}{Revisit hydrological modeling in ungauged catchments comparing regionalization, satellite observations, and machine learning approaches}.
\newblock \bibinfo{journal}{HydroResearch} .
%Type = Article
\bibitem[{Di~Baldassarre and Montanari(2009)}]{di2009uncertainty}
\bibinfo{author}{Di~Baldassarre, G.}, \bibinfo{author}{Montanari, A.}, \bibinfo{year}{2009}.
\newblock \bibinfo{title}{Uncertainty in river discharge observations: a quantitative analysis}.
\newblock \bibinfo{journal}{Hydrology and Earth System Sciences} \bibinfo{volume}{13}, \bibinfo{pages}{913--921}.
%Type = Article
\bibitem[{Eichelsd{\"o}rfer et~al.(2021)Eichelsd{\"o}rfer, Kaltenbach and Koutsourelakis}]{eichelsdorfer2021physics}
\bibinfo{author}{Eichelsd{\"o}rfer, J.}, \bibinfo{author}{Kaltenbach, S.}, \bibinfo{author}{Koutsourelakis, P.S.}, \bibinfo{year}{2021}.
\newblock \bibinfo{title}{Physics-enhanced neural networks in the small data regime}.
\newblock \bibinfo{journal}{arXiv preprint arXiv:2111.10329} .
%Type = Article
\bibitem[{Guo et~al.(2021)Guo, Leitao, Sim{\~o}es and Moosavi}]{guo2021data}
\bibinfo{author}{Guo, Z.}, \bibinfo{author}{Leitao, J.P.}, \bibinfo{author}{Sim{\~o}es, N.E.}, \bibinfo{author}{Moosavi, V.}, \bibinfo{year}{2021}.
\newblock \bibinfo{title}{Data-driven flood emulation: Speeding up urban flood predictions by deep convolutional neural networks}.
\newblock \bibinfo{journal}{Journal of Flood Risk Management} \bibinfo{volume}{14}, \bibinfo{pages}{e12684}.
%Type = Article
\bibitem[{Hao et~al.(2022)Hao, Liu, Zhang, Ying, Feng, Su and Zhu}]{hao2022physics}
\bibinfo{author}{Hao, Z.}, \bibinfo{author}{Liu, S.}, \bibinfo{author}{Zhang, Y.}, \bibinfo{author}{Ying, C.}, \bibinfo{author}{Feng, Y.}, \bibinfo{author}{Su, H.}, \bibinfo{author}{Zhu, J.}, \bibinfo{year}{2022}.
\newblock \bibinfo{title}{Physics-informed machine learning: A survey on problems, methods and applications}.
\newblock \bibinfo{journal}{arXiv preprint arXiv:2211.08064} .
%Type = Article
\bibitem[{Hrachowitz et~al.(2013)Hrachowitz, Savenije, Bl{\"o}schl, McDonnell, Sivapalan, Pomeroy, Arheimer, Blume, Clark, Ehret et~al.}]{hrachowitz2013decade}
\bibinfo{author}{Hrachowitz, M.}, \bibinfo{author}{Savenije, H.}, \bibinfo{author}{Bl{\"o}schl, G.}, \bibinfo{author}{McDonnell, J.J.}, \bibinfo{author}{Sivapalan, M.}, \bibinfo{author}{Pomeroy, J.}, \bibinfo{author}{Arheimer, B.}, \bibinfo{author}{Blume, T.}, \bibinfo{author}{Clark, M.}, \bibinfo{author}{Ehret, U.}, et~al., \bibinfo{year}{2013}.
\newblock \bibinfo{title}{A decade of predictions in ungauged basins (pub)—a review}.
\newblock \bibinfo{journal}{Hydrological Sciences Journal} \bibinfo{volume}{58}, \bibinfo{pages}{1198--1255}.
%Type = Article
\bibitem[{Jagtap et~al.(2020)Jagtap, Kharazmi and Karniadakis}]{jagtap2020conservative}
\bibinfo{author}{Jagtap, A.D.}, \bibinfo{author}{Kharazmi, E.}, \bibinfo{author}{Karniadakis, G.E.}, \bibinfo{year}{2020}.
\newblock \bibinfo{title}{Conservative physics-informed neural networks on discrete domains for conservation laws: Applications to forward and inverse problems}.
\newblock \bibinfo{journal}{Computer Methods in Applied Mechanics and Engineering} \bibinfo{volume}{365}, \bibinfo{pages}{113028}.
%Type = Article
\bibitem[{Jamali et~al.(2021)Jamali, Haghighat, Ignjatovic, Leit{\~a}o and Deletic}]{jamali2021machine}
\bibinfo{author}{Jamali, B.}, \bibinfo{author}{Haghighat, E.}, \bibinfo{author}{Ignjatovic, A.}, \bibinfo{author}{Leit{\~a}o, J.P.}, \bibinfo{author}{Deletic, A.}, \bibinfo{year}{2021}.
\newblock \bibinfo{title}{Machine learning for accelerating 2d flood models: Potential and challenges}.
\newblock \bibinfo{journal}{Hydrological Processes} \bibinfo{volume}{35}, \bibinfo{pages}{e14064}.
%Type = Article
\bibitem[{Jin et~al.(2021)Jin, Cai, Li and Karniadakis}]{jin2021nsfnets}
\bibinfo{author}{Jin, X.}, \bibinfo{author}{Cai, S.}, \bibinfo{author}{Li, H.}, \bibinfo{author}{Karniadakis, G.E.}, \bibinfo{year}{2021}.
\newblock \bibinfo{title}{Nsfnets (navier-stokes flow nets): Physics-informed neural networks for the incompressible navier-stokes equations}.
\newblock \bibinfo{journal}{Journal of Computational Physics} \bibinfo{volume}{426}, \bibinfo{pages}{109951}.
%Type = Article
\bibitem[{Kabir et~al.(2020)Kabir, Patidar, Xia, Liang, Neal and Pender}]{kabir2020deep}
\bibinfo{author}{Kabir, S.}, \bibinfo{author}{Patidar, S.}, \bibinfo{author}{Xia, X.}, \bibinfo{author}{Liang, Q.}, \bibinfo{author}{Neal, J.}, \bibinfo{author}{Pender, G.}, \bibinfo{year}{2020}.
\newblock \bibinfo{title}{A deep convolutional neural network model for rapid prediction of fluvial flood inundation}.
\newblock \bibinfo{journal}{Journal of Hydrology} \bibinfo{volume}{590}, \bibinfo{pages}{125481}.
%Type = Article
\bibitem[{Karniadakis et~al.(2021)Karniadakis, Kevrekidis, Lu, Perdikaris, Wang and Yang}]{karniadakis2021physics}
\bibinfo{author}{Karniadakis, G.E.}, \bibinfo{author}{Kevrekidis, I.G.}, \bibinfo{author}{Lu, L.}, \bibinfo{author}{Perdikaris, P.}, \bibinfo{author}{Wang, S.}, \bibinfo{author}{Yang, L.}, \bibinfo{year}{2021}.
\newblock \bibinfo{title}{Physics-informed machine learning}.
\newblock \bibinfo{journal}{Nature Reviews Physics} \bibinfo{volume}{3}, \bibinfo{pages}{422--440}.
%Type = Article
\bibitem[{Kingma and Ba(2014)}]{kingma2014adam}
\bibinfo{author}{Kingma, D.P.}, \bibinfo{author}{Ba, J.}, \bibinfo{year}{2014}.
\newblock \bibinfo{title}{Adam: A method for stochastic optimization}.
\newblock \bibinfo{journal}{arXiv preprint arXiv:1412.6980} .
%Type = Article
\bibitem[{Kratzert et~al.(2019)Kratzert, Klotz, Herrnegger, Sampson, Hochreiter and Nearing}]{kratzert2019toward}
\bibinfo{author}{Kratzert, F.}, \bibinfo{author}{Klotz, D.}, \bibinfo{author}{Herrnegger, M.}, \bibinfo{author}{Sampson, A.K.}, \bibinfo{author}{Hochreiter, S.}, \bibinfo{author}{Nearing, G.S.}, \bibinfo{year}{2019}.
\newblock \bibinfo{title}{Toward improved predictions in ungauged basins: Exploiting the power of machine learning}.
\newblock \bibinfo{journal}{Water Resources Research} \bibinfo{volume}{55}, \bibinfo{pages}{11344--11354}.
%Type = Article
\bibitem[{Kumar et~al.(2023)Kumar, Sharma, Caloiero, Mehta and Singh}]{kumar2023comprehensive}
\bibinfo{author}{Kumar, V.}, \bibinfo{author}{Sharma, K.V.}, \bibinfo{author}{Caloiero, T.}, \bibinfo{author}{Mehta, D.J.}, \bibinfo{author}{Singh, K.}, \bibinfo{year}{2023}.
\newblock \bibinfo{title}{Comprehensive overview of flood modeling approaches: A review of recent advances}.
\newblock \bibinfo{journal}{Hydrology} \bibinfo{volume}{10}, \bibinfo{pages}{141}.
%Type = Article
\bibitem[{LeCun et~al.(2015)LeCun, Bengio and Hinton}]{lecun2015deep}
\bibinfo{author}{LeCun, Y.}, \bibinfo{author}{Bengio, Y.}, \bibinfo{author}{Hinton, G.}, \bibinfo{year}{2015}.
\newblock \bibinfo{title}{Deep learning}.
\newblock \bibinfo{journal}{Nature} \bibinfo{volume}{521}, \bibinfo{pages}{436--444}.
%Type = Article
\bibitem[{L{\"o}we et~al.(2021)L{\"o}we, B{\"o}hm, Jensen, Leandro and Rasmussen}]{lowe2021u}
\bibinfo{author}{L{\"o}we, R.}, \bibinfo{author}{B{\"o}hm, J.}, \bibinfo{author}{Jensen, D.G.}, \bibinfo{author}{Leandro, J.}, \bibinfo{author}{Rasmussen, S.H.}, \bibinfo{year}{2021}.
\newblock \bibinfo{title}{U-flood--topographic deep learning for predicting urban pluvial flood water depth}.
\newblock \bibinfo{journal}{Journal of Hydrology} \bibinfo{volume}{603}, \bibinfo{pages}{126898}.
%Type = Article
\bibitem[{Lu et~al.(2021)Lu, Meng, Mao and Karniadakis}]{lu2021deepxde}
\bibinfo{author}{Lu, L.}, \bibinfo{author}{Meng, X.}, \bibinfo{author}{Mao, Z.}, \bibinfo{author}{Karniadakis, G.E.}, \bibinfo{year}{2021}.
\newblock \bibinfo{title}{Deepxde: A deep learning library for solving differential equations}.
\newblock \bibinfo{journal}{SIAM review} \bibinfo{volume}{63}, \bibinfo{pages}{208--228}.
%Type = Inproceedings
\bibitem[{Mahesh et~al.(2022)Mahesh, Leandro and Lin}]{mahesh2022physics}
\bibinfo{author}{Mahesh, R.B.}, \bibinfo{author}{Leandro, J.}, \bibinfo{author}{Lin, Q.}, \bibinfo{year}{2022}.
\newblock \bibinfo{title}{Physics informed neural network for spatial-temporal flood forecasting}, in: \bibinfo{booktitle}{Climate Change and Water Security: Select Proceedings of VCDRR 2021}, \bibinfo{organization}{Springer}. pp. \bibinfo{pages}{77--91}.
%Type = Inproceedings
\bibitem[{Nair and Hinton(2010)}]{nair2010rectified}
\bibinfo{author}{Nair, V.}, \bibinfo{author}{Hinton, G.E.}, \bibinfo{year}{2010}.
\newblock \bibinfo{title}{Rectified linear units improve restricted boltzmann machines}, in: \bibinfo{booktitle}{Proceedings of the 27th international conference on machine learning (ICML-10)}, pp. \bibinfo{pages}{807--814}.
%Type = Inproceedings
\bibitem[{Ng(2004)}]{ng2004feature}
\bibinfo{author}{Ng, A.Y.}, \bibinfo{year}{2004}.
\newblock \bibinfo{title}{Feature selection, l 1 vs. l 2 regularization, and rotational invariance}, in: \bibinfo{booktitle}{Proceedings of the twenty-first international conference on Machine learning}, p.~\bibinfo{pages}{78}.
%Type = Article
\bibitem[{Nguyen et~al.(2023)Nguyen, Dang, Nguyen, Pham~Van, Van~Nguyen, Nguyen, Nguyen, Pham, Pham and Bui}]{nguyen2023integration}
\bibinfo{author}{Nguyen, H.D.}, \bibinfo{author}{Dang, D.K.}, \bibinfo{author}{Nguyen, Y.N.}, \bibinfo{author}{Pham~Van, C.}, \bibinfo{author}{Van~Nguyen, T.T.}, \bibinfo{author}{Nguyen, Q.H.}, \bibinfo{author}{Nguyen, X.L.}, \bibinfo{author}{Pham, L.T.}, \bibinfo{author}{Pham, V.T.}, \bibinfo{author}{Bui, Q.T.}, \bibinfo{year}{2023}.
\newblock \bibinfo{title}{Integration of machine learning and hydrodynamic modeling to solve the extrapolation problem in flood depth estimation}.
\newblock \bibinfo{journal}{Journal of Water and Climate Change} , \bibinfo{pages}{jwc2023573}.
%Type = Article
\bibitem[{Prestininzi et~al.(2011)Prestininzi, Di~Baldassarre, Schumann and Bates}]{prestininzi2011selecting}
\bibinfo{author}{Prestininzi, P.}, \bibinfo{author}{Di~Baldassarre, G.}, \bibinfo{author}{Schumann, G.}, \bibinfo{author}{Bates, P.}, \bibinfo{year}{2011}.
\newblock \bibinfo{title}{Selecting the appropriate hydraulic model structure using low-resolution satellite imagery}.
\newblock \bibinfo{journal}{Advances in Water Resources} \bibinfo{volume}{34}, \bibinfo{pages}{38--46}.
%Type = Article
\bibitem[{Qian et~al.(2019)Qian, Mohamed and Claudel}]{qian2019physics}
\bibinfo{author}{Qian, K.}, \bibinfo{author}{Mohamed, A.}, \bibinfo{author}{Claudel, C.}, \bibinfo{year}{2019}.
\newblock \bibinfo{title}{Physics informed data driven model for flood prediction: application of deep learning in prediction of urban flood development}.
\newblock \bibinfo{journal}{arXiv preprint arXiv:1908.10312} .
%Type = Article
\bibitem[{Raissi et~al.(2019)Raissi, Perdikaris and Karniadakis}]{raissi2019physics}
\bibinfo{author}{Raissi, M.}, \bibinfo{author}{Perdikaris, P.}, \bibinfo{author}{Karniadakis, G.E.}, \bibinfo{year}{2019}.
\newblock \bibinfo{title}{Physics-informed neural networks: A deep learning framework for solving forward and inverse problems involving nonlinear partial differential equations}.
\newblock \bibinfo{journal}{Journal of Computational Physics} \bibinfo{volume}{378}, \bibinfo{pages}{686--707}.
%Type = Article
\bibitem[{Rumelhart et~al.(1986)Rumelhart, Hinton and Williams}]{rumelhart1986learning}
\bibinfo{author}{Rumelhart, D.E.}, \bibinfo{author}{Hinton, G.E.}, \bibinfo{author}{Williams, R.J.}, \bibinfo{year}{1986}.
\newblock \bibinfo{title}{Learning representations by back-propagating errors}.
\newblock \bibinfo{journal}{Nature} \bibinfo{volume}{323}, \bibinfo{pages}{533--536}.
%Type = Article
\bibitem[{Shorten and Khoshgoftaar(2019)}]{shorten2019survey}
\bibinfo{author}{Shorten, C.}, \bibinfo{author}{Khoshgoftaar, T.M.}, \bibinfo{year}{2019}.
\newblock \bibinfo{title}{A survey on image data augmentation for deep learning}.
\newblock \bibinfo{journal}{Journal of Big Data} \bibinfo{volume}{6}, \bibinfo{pages}{1--48}.
%Type = Article
\bibitem[{Zhu et~al.(2019)Zhu, Zabaras, Koutsourelakis and Perdikaris}]{zhu2019physics}
\bibinfo{author}{Zhu, Y.}, \bibinfo{author}{Zabaras, N.}, \bibinfo{author}{Koutsourelakis, P.S.}, \bibinfo{author}{Perdikaris, P.}, \bibinfo{year}{2019}.
\newblock \bibinfo{title}{Physics-constrained deep learning for high-dimensional surrogate modeling and uncertainty quantification without labeled data}.
\newblock \bibinfo{journal}{Journal of Computational Physics} \bibinfo{volume}{394}, \bibinfo{pages}{56--81}.

\end{thebibliography}



%% For example:

%% \bibitem[Aladro et al.(2015)]{Aladro15} Aladro, R., Martín, S., Riquelme, D., et al. 2015, \aas, 579, A101

%%\end{thebibliography}

\end{document}